ns-

\documentclass[11pt, letter, onecolumn]{arxiv}
\usepackage{dm-colors}
\usepackage{graphicx}%
\usepackage{epsfig}
\usepackage{multirow}%
\usepackage{amsmath,amssymb,amsfonts}%
\usepackage{amsthm}%
\usepackage{mathrsfs}%
\usepackage[title]{appendix}%
\usepackage{xcolor}%
\usepackage{textcomp}%
\usepackage{manyfoot}%
\usepackage{booktabs}%
\usepackage{algorithm}%
\usepackage{algorithmicx}%
\usepackage{algpseudocode}%
\usepackage{listings}%
\usepackage{caption}
\usepackage{subcaption}
\usepackage[inkscapelatex=false]{svg}
\usepackage{multirow}
\usepackage{tikz}
\usepackage{hyperref}

\newcommand\PlaceText[3]{%
\begin{tikzpicture}[remember picture,overlay]
\node[outer sep=0pt,inner sep=0pt,anchor=south west] 
  at ([xshift=#1,yshift=-#2]current page.north west) {#3};
\end{tikzpicture}%
}

\usepackage{comment}
\usepackage{bbding}
\usepackage[bibstyle=nature,citestyle=numeric-comp,%
            natbib=true,backend=biber,maxbibnames=99,%
            giveninits=false,sorting=none]{biblatex}
\graphicspath{ {Sections/Figures} }

\title{{EchoApex: A General-Purpose Vision Foundation Model for Echocardiography}}
\author[$\ast$]{Abdoul Aziz Amadou}
\author[$\ast,\dagger$]{Yue Zhang}
\author[\empty{}]{S\'ebastien Piat}
\author[\empty{}]{Paul Klein}
\author[\empty{}]{Ingo Schmuecking}
\author[\empty{}]{Tiziano Passerini}
\author[$\dagger$]{Puneet Sharma} 
\affil[\empty{}]{Digital Technology \& Innovation, Siemens Healthineers}
\renewcommand{\correspondingauthor}[1]{$\ast$~Equal contributions. %
                                       $\dagger$~Corresponding authors: \{yue.zhang, sharma.puneet\}@siemens-healthineers.com \\
                                       \textbf{Disclaimer}: The concepts and information presented in this paper/presentation are based on research results that are not commercially available. Future commercial availability cannot be guaranteed.}
                                       
\begin{abstract}
\textbf{Purpose:}  Quantitative evaluation of echocardiography is essential for precise assessment of cardiac condition, monitoring disease progression, and guiding treatment decisions. The diverse nature of echo images, including variations in transducer types, manufacturers, and pathologies, poses challenges for developing artificial intelligent models that can generalize across different clinical practice. We introduce EchoApex, a general-purpose vision foundation model for a variety of clinical applications in echocardiography.

\textbf{Methods:}  Leveraging self-supervised learning, EchoApex is pretrained on over 20 million echo images from 11 clinical centres. By incorporating task-specific decoders and adapter modules, we demonstrate the effectiveness of EchoApex on 4 different types of clinical applications with 28 sub-tasks. Specifically, we evaluate the pretrained EchoApex encoding model on: 1) sequence view classification with linear decoder on 18 different view points; 2) interactive structure segmentation with prompt-based encoder-decoder on chambers from different views; 3) left ventricle hypertrophy detection with multiscale convolutional decoder, and 4) ejection fraction estimation with a spatial-temporal sequence decoder. In addition, we investigate the learning efficiency and generalizability of EchoApex via different learning settings, including supervised fine-tuning, zero and few-shot learning. Finally, we study the feasibility of parameter-efficient adaptation of EchoApex to aforementioned downstream tasks with limited computational resources.

\textbf{Results:} Compared to state-of-the-art task-specific models, EchoApex attains improved performance on all evaluated applications with a unified image encoding architecture. For the view classification task, studied on over 26K annotated videos, EchoApex reaches a mean BACC of 0.976, showing improvement ($p<0.01$) on 14 out of 18 view points over the ImageNet pretrained counterpart, with an average increment of 0.02 BACC. For the segmentation task, benchmarked on the 3 largest publicly available echocardiogram datasets with a total number of 74K annotated frames, EchoApex shows an improved DICE (0.927, 95\% CI 0.926-0.928) compared to sub-task specialist models (e.g. UNet and DeepLabV3, 0.904) as well as generalist model MedSAM (0.870, 95\% CI 0.868-0.871) on all targets. In a zero-shot test, EchoApex shows improvement (0.877, 95\% CI 0.873-0.880) over specialist models (0.834, 95\% CI 0.829-0.840) in all evaluated datasets. On the left ventricle measurement task, studied on 12K annotated videos, EchoApex shows an improved MAE of 0.2 mm compared with the specialist baseline, and a significant improvement of 1.4mm on 1K frames from a cohort of unseen data, demonstrating its generalization capability. For ejection fraction estimation, evaluated on a dataset with over 1.2K patients and with a frozen encoder, EchoApex achieved MAE of 5.6\% and AUC of 0.93 for cardiomyopathy detection with a simple spatial-temporal model extension, showing improvement compared to 6.1\% of MAE and 0.89 of AUC from the ImageNet pretrained model. With the pretrained encoder frozen, EchoApex with adapters uses less than 4\% of trainable parameters but showed competitive performance with a maximum of 5.1\% degradation from fully finetuned models.

\textbf{Conclusion:} The improved performance of EchoApex compared with existing works demonstrated the benefits of model pretraining at scale with in-domain data. Furthermore, EchoApex illustrates the potential for developing a general-purpose vision foundation model tailored specifically for echocardiography, capable of addressing a diverse range of clinical applications with high efficiency and efficacy.
\end{abstract}

\begin{document}
\maketitle

\section{Introduction}\label{sec1:introduction}
Echocardiography is deeply integrated into a wide range of clinical practice, covering diagnosis, intervention and follow-up of cardiovascular disease \cite{nolan2019automated, rwebembera20242023, little2023recommendations}. The increasing usage of echocardiography and the diverse range of tasks bring new challenges to clinicians, who are often required to interpret numerous images in a short amount of time \cite{onwordi2024prevalence, engen2010effect}. This has led to a growing demand for automated analysis tools to support clinicians in their daily tasks. In recent years, the field of artificial intelligence has witnessed the emergence of a novel category of models known as Foundation Models. These models, primarily developed for general computer vision, are trained on extensive datasets of natural images to capture the intrinsic properties of a diverse array of data \cite{chen2020simple, caron2020unsupervised, caron2021emerging, he2022masked, oquab2023dinov2}. The overarching aim is to create a single, robust model that can subsequently be fine-tuned for a variety of specific tasks, thus significantly enhancing its adaptability and performance across different applications. This paradigm shift has led to remarkable advancements, particularly in clinical modalities such as pathology and X-ray imaging, where these models have demonstrated significant improvements in performance and accuracy \cite{moor2023foundation, he2024foundation, chen2024towards, vorontsov2023virchow, ghesu2022contrastive}. However, the potential of foundation models in echocardiography remains largely unexplored.

The benefits of training such a foundational model for echocardiography has been partially revealed in the development of segmentation models. Trained on tens of thousands of echocardiographic images,  recent models, such as SAMUS \cite{lin2023samus} and SonoSAM \cite{ravishankar2023sonosam}, have been developed with a focus on specific tasks like segmentation. These models have shown promising performance within their respective domains, underscoring the feasibility and utility of foundation models in echocardiography. However, the task-specific nature of these models limits their broader applicability.  Concurrently, advances in vision-language models for echocardiography \cite{christensen2024vision,vukadinovic2024echoprime} have shown the potential of using video-report pairs to enhance image interpretation. In contrast, a vision foundation model that focuses on a different set of applications and does not rely on report input can serve as a versatile tool across a wide range of clinical workflows. Its primary focus is to automate processes and improve diagnostic accuracy in echocardiography without the need for textual data inputs.

The creation of a general-purpose vision foundation model for echocardiography involves several challenges. First, the collection and curation of a sufficiently large and diverse dataset are critical. This dataset must encompass a wide range of imaging scenarios and hardware specs to ensure the model's robustness and generalizability. Second, the model must be designed to effectively capture the extensive nature of echocardiographic data. As a live, cost-effective imaging modality, echocardiography is much more accessible compared to computed tomography (CT) and magnetic resonance imaging (MRI). Consequently, echocardiographic images are continuously acquired and quickly accumulate into a large volume. Furthermore, the model must be capable of fine-tuning for specific tasks while maintaining its overall adaptability. This involves developing efficient algorithms for transfer learning and task-specific adaptation. 

Building on prior efforts, we propose a general-purpose vision foundation model for echocardiography, trained on a large-scale dataset of 20 million echocardiographic images derived from 450,338 videos involving 26,000 patients across 11 clinical centers. This dataset, termed Echo-20M, encompasses a comprehensive range of echocardiographic data, including transthoracic echocardiograms (TTE) from diagnostic exams, as well as transesophageal echocardiograms (TEE) and intracardiac echocardiograms (ICE) obtained from all major types of echo-guided cardiac interventions. To closely mirror the diverse nature of clinical echocardiography, Echo-20M comprises B-mode images, Doppler images, and volumetric images. During the pretraining stage, we employ state-of-the-art self-supervised learning method DINOv2 \cite{oquab2023dinov2}, which has demonstrated promising results when adapted to the medical domain \cite{chen2024towards, xu2024whole, song2024general}. This approach enables our model to learn robust representations from the vast and varied dataset. In the fine-tuning stage, we showcase the versatility of our model by attaching task-specific heads to perform a wide array of applications in cardiac care. These applications include a linear layer for view classification, a two-way transformer decoder for interactive segmentation \cite{kirillov2023segment}, a sequence decoder for ejection fraction computation, and a multi-scale convolutional decoder for left ventricular hypertrophy (LVH) detection \cite{hatamizadeh2022unetr}. Our findings indicate that this model achieves state-of-the-art performance across these diverse tasks with a single, unified model trained on the extensive Echo-20M dataset. Furthermore, by leveraging the latest practices with adapters \cite{chen2022vision}, we demonstrate that our model can be effortlessly adapted to downstream tasks with minimal additional training. This adaptability underscores the model's potential as a versatile and efficient tool for a wide range of clinical applications, paving the way for enhanced diagnostic accuracy and improved patient outcomes in echocardiography.

\begin{figure*}
\centering
    \begin{subfigure}{\textwidth}
        \centering
         \includegraphics[width=\textwidth]{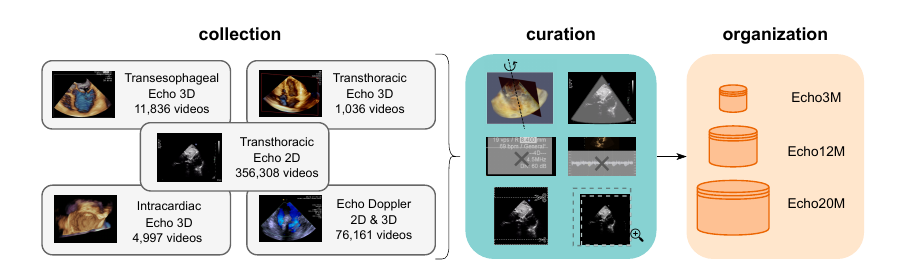} 
        \label{fig:ssl_pipeline}
         \vspace{-0.2in}
     \end{subfigure}
     \PlaceText{21mm}{33mm}{\textbf{(a)}}
     \begin{subfigure}{0.6\textwidth}
         \centering
         \includegraphics[width=\textwidth]{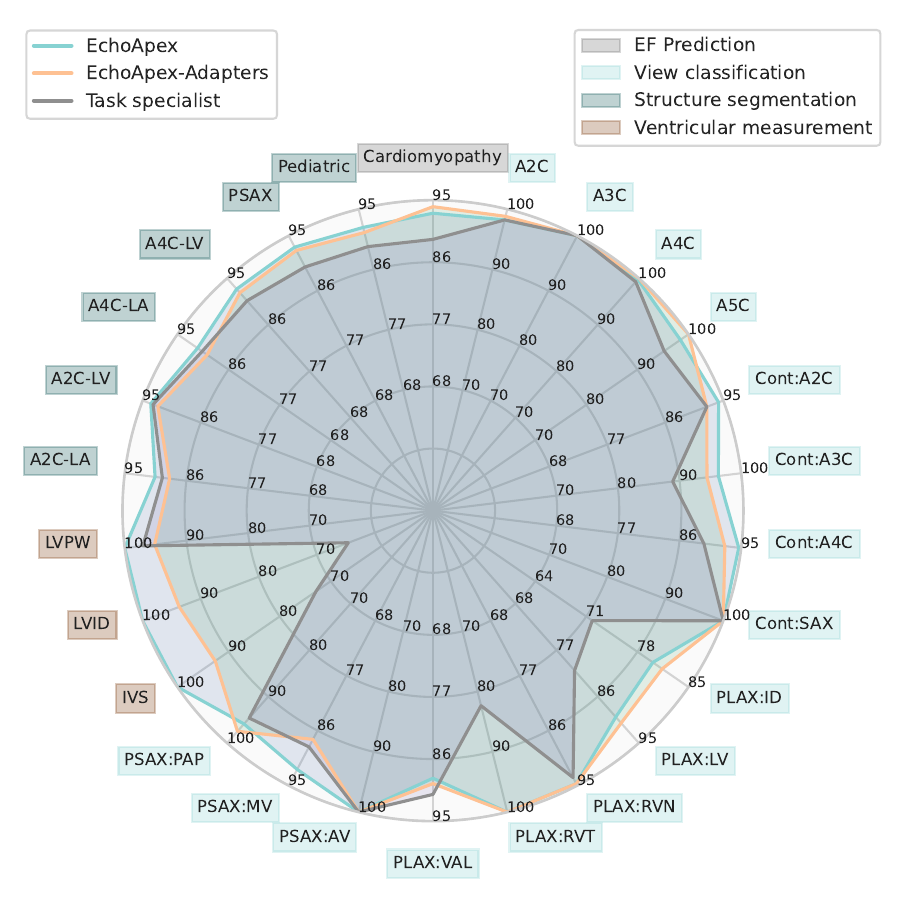}
         \label{fig:ssl_results}
     \end{subfigure}
     \PlaceText{21mm}{86mm}{\textbf{(b)}}
     \hfill
     \begin{subfigure}{0.39\textwidth}
         \centering
         \includegraphics[width=\textwidth]{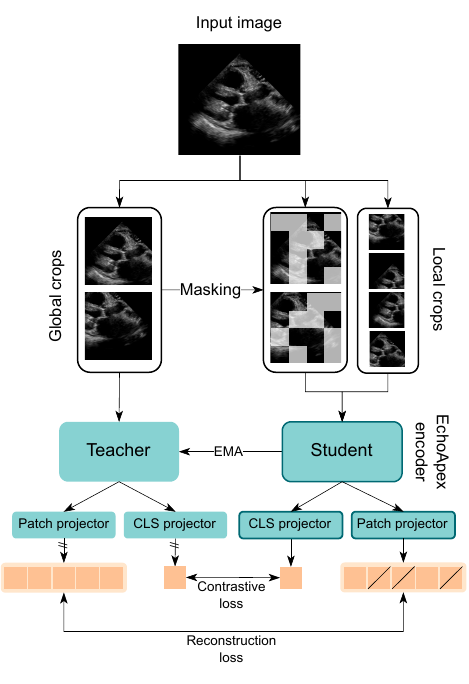}
         \vspace{0.1in}
         \label{fig:ssl_pretrain}
     \end{subfigure}
     \begin{subfigure}{1.01\textwidth}
         \centering
         \vspace{-0.2in}
         \hspace{-0.5in}
         \includegraphics[width=\textwidth]{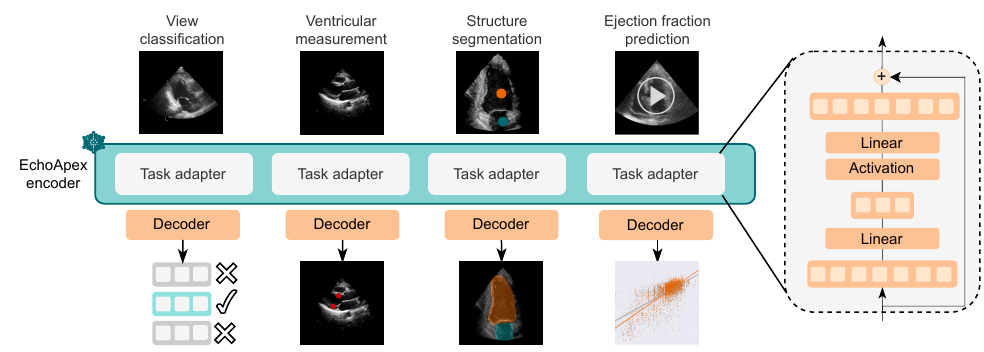}
         \label{fig:echoapex}
     \end{subfigure}
    \PlaceText{130mm}{86mm}{\textbf{(c)}}
    \PlaceText{21mm}{185mm}{\textbf{(d)}}
    \caption{\textbf{Overview of EchoApex.} EchoApex is a general-purpose vision foundation model enpowering a diverse range of clinical tasks in echocardiography. \textbf{(a)} Data curation process. A total number of 450K videos are collected from 11 clinical centres, covering different image characteristics. \textbf{(b)} Evaluation of EchoApex on view classification, structure segmentation, ventricular measurement and automated EF prediction. EchoApex shows superior performance to the task specialist in all evaluated tasks. \textbf{(c)} EchoApex pretraining with the state-of-the-art self-supervised training algorithm DINOv2. \textbf{(d)} EchoApex application in  downstream tasks with optional parameter efficient model fine-tuning. }
    \label{fig:main_figure}
\end{figure*}

\section{Results}
\subsection{Pretraining}\label{sec2:pretraining}

A defining feature of foundation models is their remarkable ability to improve performance on downstream tasks as they scale with increased data and model size. This scalability has been demonstrated across various domains, including natural image processing\cite{oquab2023dinov2}, computational pathology\cite{vorontsov2023virchow, chen2024towards}, and medical imaging such as X-rays\cite{ghesu2022contrastive}. In this study, we investigate whether this trend extends to model pretraining in echocardiography. We perform a series of experiments varying both the dataset size and the model architecture to systematically evaluate their impact on performance.

To evaluate the performance of different neural network architectures, we pretrain three distinct models: Vision Transformer Small (ViT-S)\cite{dosovitskiy2020image}, Vision Transformer Base (ViT-B), and ResNet50\cite{He2015ResNet}. To assess the impact of dataset scale and characteristics, we create three subsets: 1) 3 million biplane images (Echo3M) derived from echocardiogram acquired using volume transducers, 2) 12 million images (Echo12M) randomly sampled from the entire dataset, and 3) 20 million images (Echo20M), an enhanced subset of Echo12M encompassing a comprehensive range of image types at a larger scale. Models on Echo3M and Echo12M are pretrained from scratch, while models on Echo20M continue pretraining from the Echo12M models. 

We utilize K-Nearest Neighbors (KNN) on a view classification task to evaluate the effectiveness of the pretrained embeddings in distinguishing between different echocardiographic views. Our results indicate that when pretrained on Echo3M, the ViT-S model achieves only 70.2\% accuracy, with minimal improvement over the entire training process. This is likely due to Echo3M containing only biplanes from volume transducers, which differ from the transducers used in 2D echocardiograms, resulting in poor performance on images with unseen characteristics. This highlights the importance of a diverse dataset for pretraining. In contrast, the same model pretrained on Echo12M achieves 87.5\% accuracy, with performance steadily increasing throughout the training process and further improving to a peak of 91.5\% when pretrained on Echo20M. The ResNet50, of similar size to ViT-S, attains a lower peak performance of 76.6\% on Echo12M and 88.4\% on Echo20M. The larger ViT-B model pretrained on Echo12M attains 90.5\% accuracy and reaches a maximum of 91.9\% when pretrained on Echo20M. The performance gain for ViT-B is also more stable compared to ViT-S during pretraining.  These observations are consistent with previous studies \cite{vorontsov2023virchow, chen2024towards}, confirming that model performance improves with increased dataset size and model complexity. We further analyzed the ViT-B feature embeddings using a T-SNE plot, which shows well-separated clusters for different view classes, indicating a strong capability for unsupervised learning from the data. Additional details regarding the experimental setup, implementation, and performance metrics are provided in the Methods section.

\begin{figure}[H]
  \begin{subfigure}[b]{0.29\linewidth}
    \centering
    \includegraphics[width=0.93\linewidth]{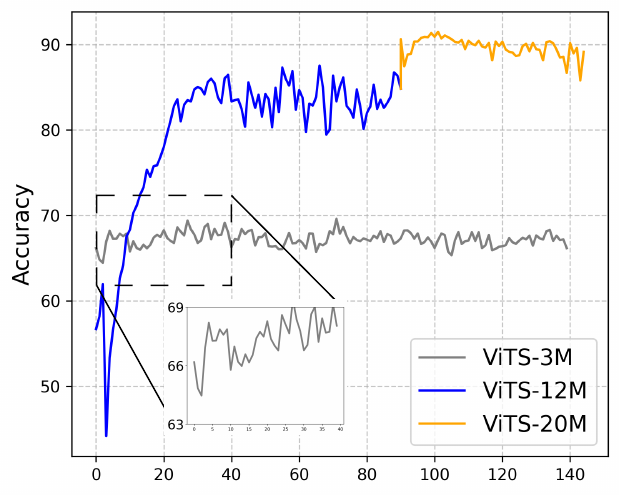}
    \includegraphics[width=0.93\linewidth]{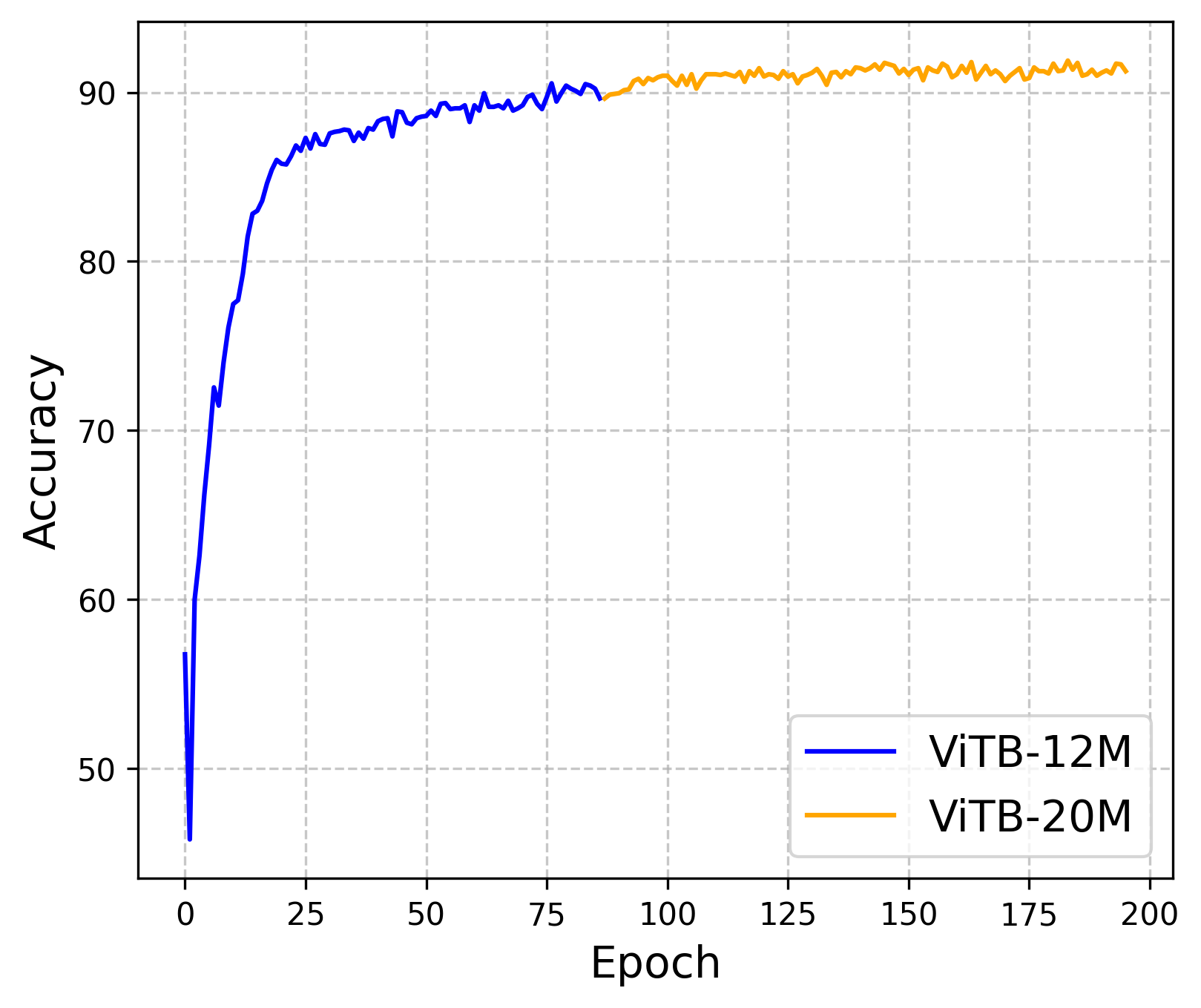}  
  \end{subfigure}
  \begin{subfigure}{0.71\linewidth}
    \centering
    \includegraphics[width=\linewidth]{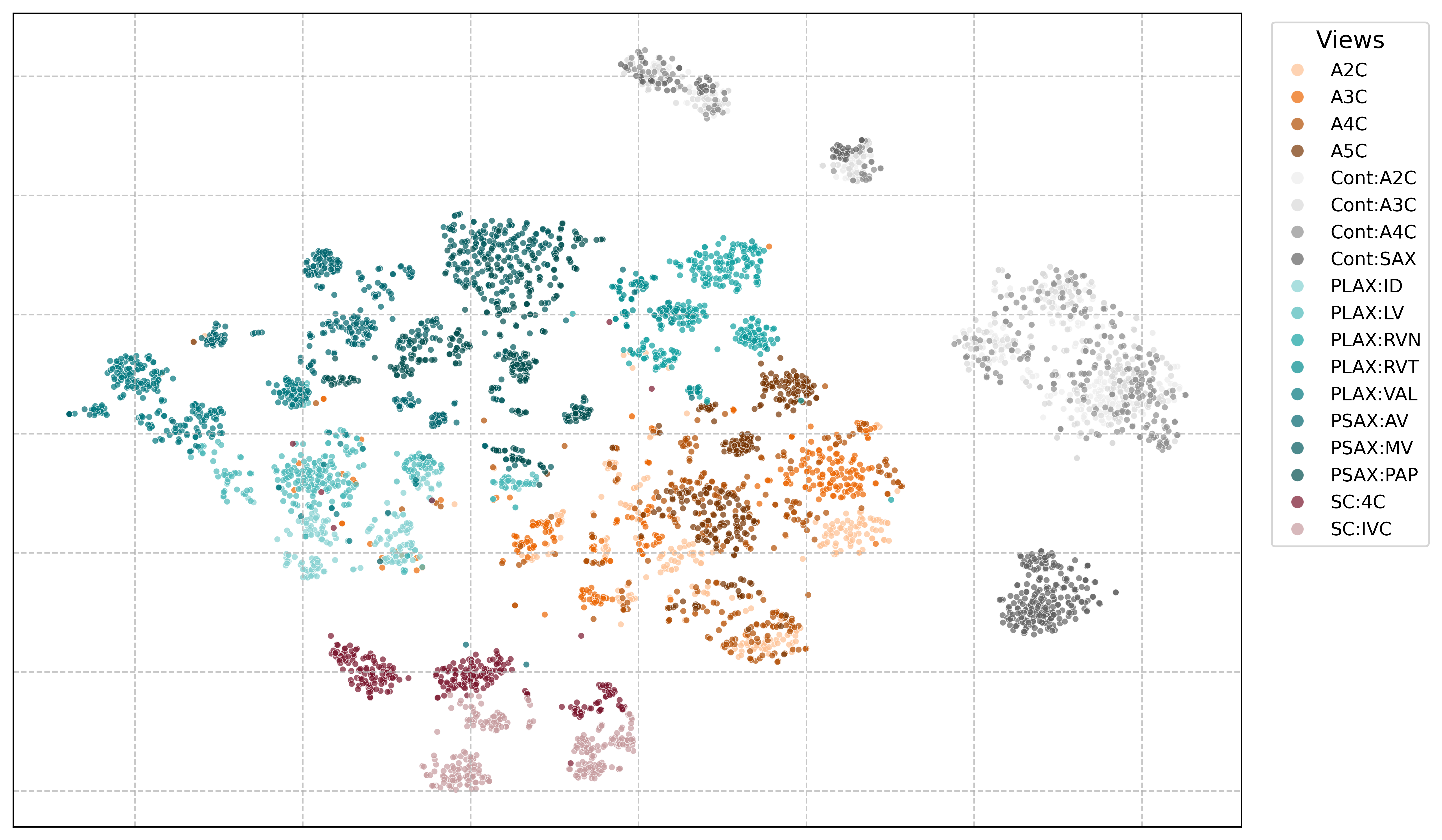}
    \vspace{0.002in}
  \end{subfigure}
  \caption{Qualitative and quantitative evaluation of model pretraining. Left top: KNN accuracy at different training epochs during the pretraining of ViT-S model. Right bottom: KNN accuracy at different training epochs during pretraining of ViT-B model. For both plots, blue segment represents pretraining on Echo12M and orange segment represents continual training on echo20M.  Right: Two-dimensional t-SNE visualization of the embeddings of 10K images. The  encoder is ViT-B pretrained on Echo20M. A perplexity of 50 is used. }
\end{figure}

\subsection{View Classification}\label{sec3:viewclassification}

TTE is widely used in cardiology for the diagnosis and follow up of patients affected by cardiovascular disease. Echo studies obtained from TTE examinations cover a wide range of views, resulting in large datasets to be reviewed by clinicians. AI-based models for TTE view classification can prove beneficial in identifying clips of desired views or generating structured reports. For this task, we use an internal multi-vendor dataset, including a wide range of transducers, spatial and temporal resolutions, image quality, imaging depth, color doppler and the use of contrast. A total of 18 different TTE views are covered in the datasets, including standard views and those where contrast is used. All the images were labeled by echocardiographers following the recommended guidelines \cite{Mitchell2019GuidelinesFP}.

\begin{figure*}
     \centering
     \begin{subfigure}{0.35\textwidth}
         \centering
         \includegraphics[width=0.9\textwidth]{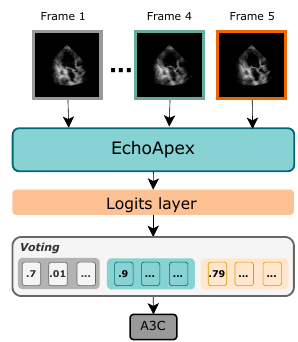}
        \captionlistentry{}
        \label{fig:view_cls_diagram}
     \end{subfigure}
     \PlaceText{25mm}{37mm}{\textbf{(a)}}
     \hfill
     \begin{subfigure}{0.59\textwidth}
         \centering
         \includegraphics[width=0.9\textwidth]{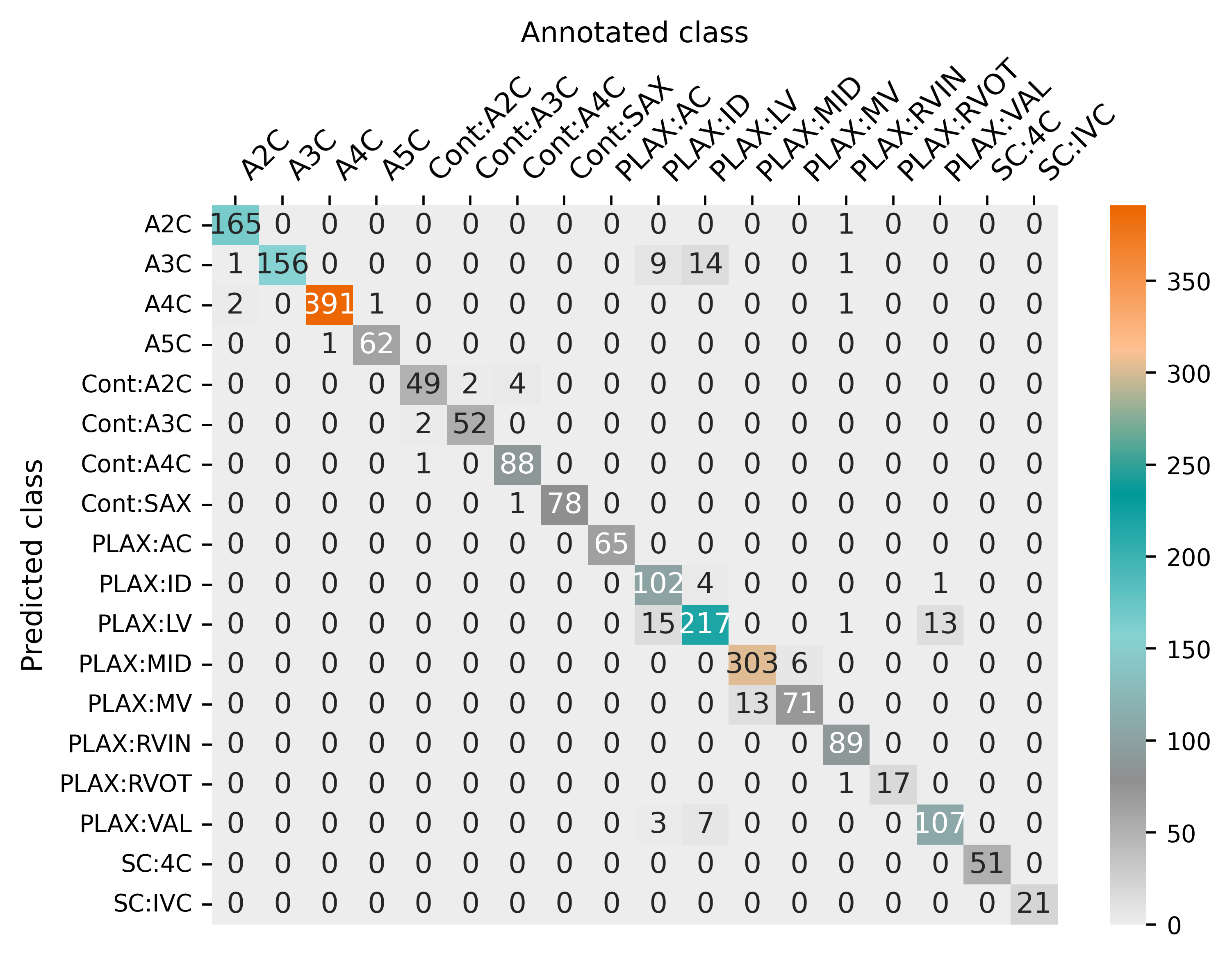}
         \captionlistentry{}
         \label{fig:view_cls_confusion}
     \end{subfigure}
     \PlaceText{100mm}{37mm}{\textbf{(b)}}
     \hfill
     \vspace{0.2in}
     \begin{subfigure}{1.0\textwidth}
         \centering
         \includegraphics[width=\textwidth]{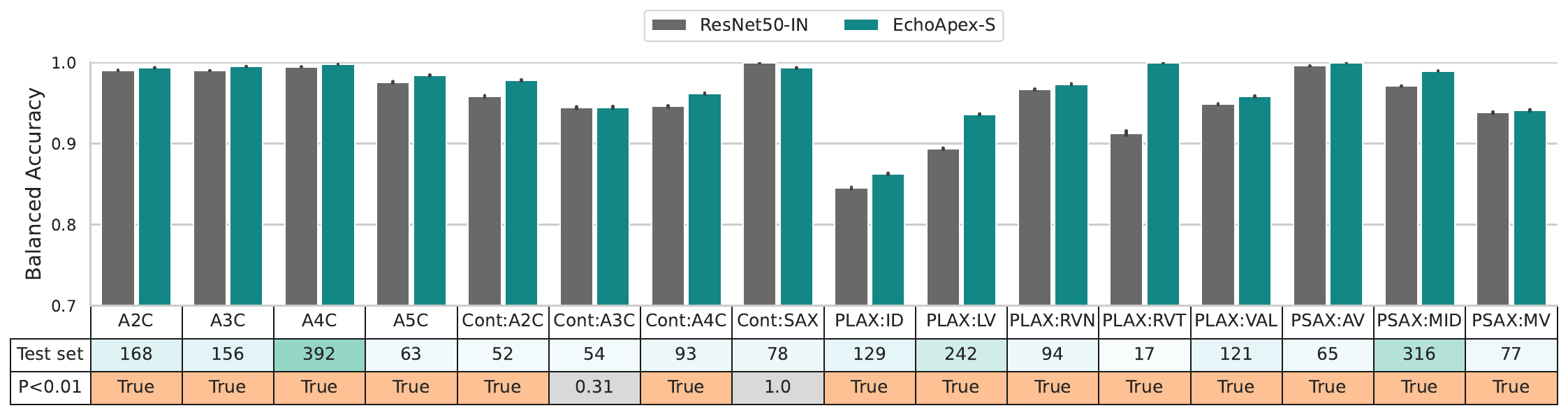}
         \captionlistentry{}
         \label{fig:view_cls_bar_and_table}
     \end{subfigure}
     \PlaceText{25mm}{103mm}{\textbf{(c)}}
     \hfill
     \begin{subfigure}[b]{1.0\textwidth}
         \centering
         \includegraphics[width=0.95\textwidth]{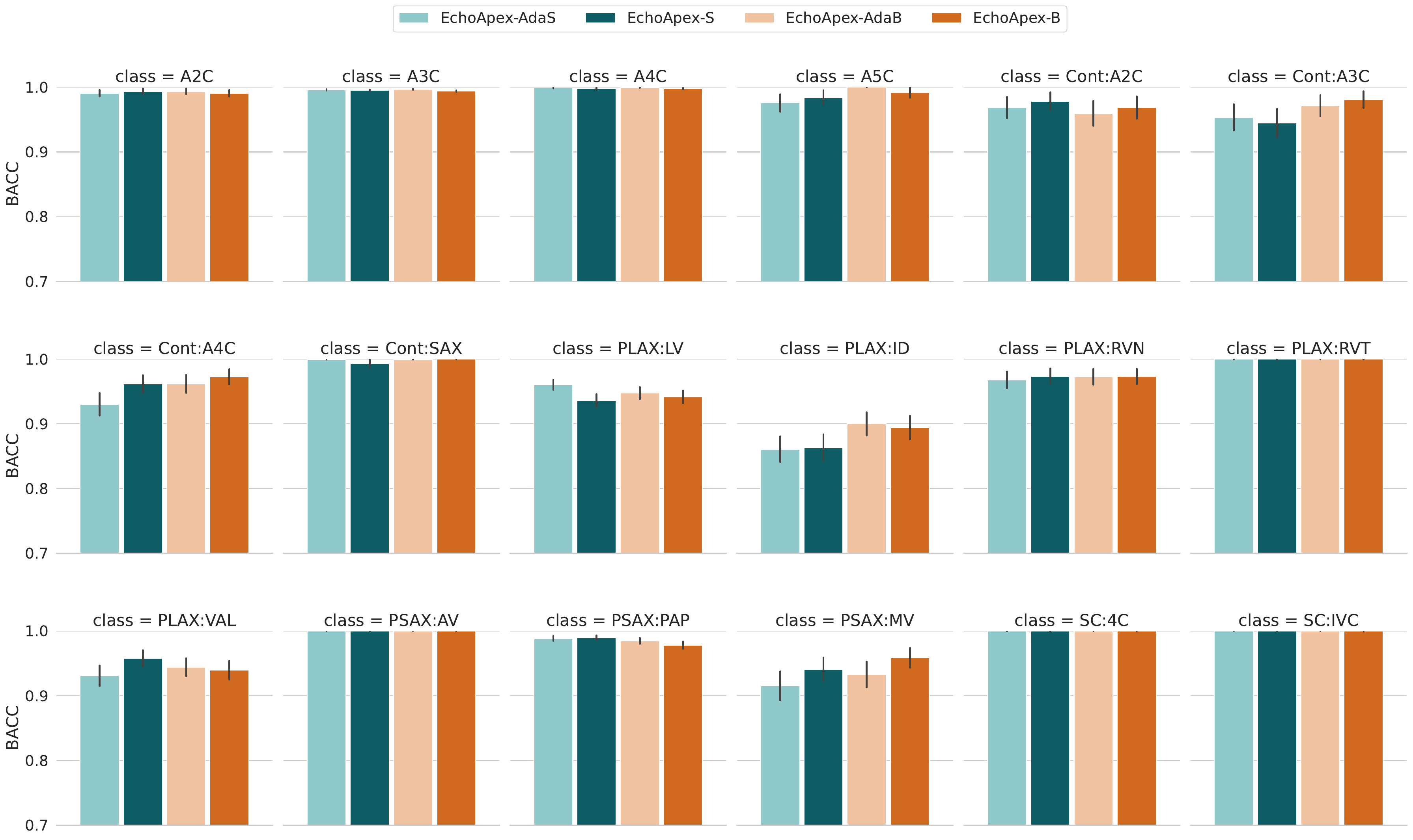}
         \captionlistentry{}
         \label{fig:view_cls_gridplot}
     \end{subfigure}
     \PlaceText{25mm}{156mm}{\textbf{(d)}}
        \caption{\textbf{Study on sequence view classification.} \textbf{(a)} Architecture for the classification task. \textbf{(b)} Confusion matrix showing EchoApex-B's performance over the 18 classes. \textbf{(c)} Performance comparison of EchoApex-S and the ResNet baseline. The table indicates the number of test sequences and whether or not EchoApex-S significantly outperforms the baseline. \textbf{(d)} Balanced accuracy (BACC) of the EchoApex models trained with different backbones and with(out) adapters.}
    \label{fig:view_cls_main_figure}
\end{figure*}

\subsubsection{Supervised view classification with linear classifier}
The model architecture for this task is based on our pretrained EchoApex backbone and a linear classification layer, which uses the features from the classification (CLS) token to produce the logits, following Fig. \ref{fig:view_cls_diagram}. We use a weighted cross-entropy loss to account for class imbalance. The distribution of the classes in our dataset, sample images from the various classes and the list of hyperparameters for this experiment are given in the Supplementary Fig. \ref{fig:supp_view_cls_distribution}, \ref{fig:supp_view_cls_gallery} and Table. \ref{table:supp_view_cls_hyperparams}, respectively.

When testing our model, we extract 5 frames from each video sequence, classify them individually and apply majority voting to obtain a sequence label. We found this strategy to be more effective in reducing confusion. Indeed, using multiple frames can help with ambiguity between views due to the cardiac motion during the cardiac cycle, e.g. between the apical four and five chamber views. The confusion matrix in Fig. \ref{fig:view_cls_confusion} details the results of the best performing EchoApex model with a ViT-B backbone (EchoApex-B) over the 18 classes. The confusion between the parasternal long axis left ventricle (PLAX LV), increased depth (PLAX ID), and zoomed on the mitral and/or aortic valves (PLAX Valves) is due to the same structures being present in the three views, with the main difference between them being the varying zoom levels. Finally, the A3C and the PLAX ID/LV views exhibit the same anatomical structures but the angulation differs between the two sets of views. The model tends to make errors when the image quality is poor or the angulation not adequate.

For all the subsequent experiments, we report the Balanced Accuracy (BACC) of the models, which takes in account the class imbalance in the dataset. We first compare EchoApex-S with a baseline ResNet50 \cite{He2015ResNet} pretrained on ImageNet-1K \cite{Deng2009ImageNetAL} (ResNet50-IN) in Fig. \ref{fig:view_cls_bar_and_table}. Our model significantly outperforms the baseline on 14 of the 18 subtasks ($P < 0.01$, according to a one-sided permutation test). We omit the subcostal four chamber and inferior vena cava views (SC:4C and SC:IVC) in Fig. \ref{fig:view_cls_bar_and_table} as both models achieve mean BACC values of 0.99. EchoApex-S shows stronger performance on the less prevalent parasternal classes, with the most notable improvements in the PLAX:RVT and PLAX:LV classes, where we report mean BACC values of 0.99 and 0.94 respectively, as opposed to values of 0.91 and 0.89 for the ResNet50-IN baseline. Those results highlight the benefits of pretraining on a large-scale echo database.

In Fig. \ref{fig:view_cls_bar_and_table}, when comparing versions of the EchoApex model varying with model size (Vit-S vs Vit-B) and training strategy (fine-tuned vs adapters), the fine-tuned EchoApex-B model achieves the highest mean BACC (0.976) over all the classes, compared to values of 0.972 for EchoApex-S, 0.975 for EchoApex-B with a ViT-B backbone and adapters (EchoApex-AdaB), and 0.968 for EchoApex-AdaS. When comparing the fine-tuned models, EchoApex-B outperforms EchoApex-S significantly on 8 out of the 18 classes ($P < 0.01$, no significant difference in performance on 4 classes), mostly on the contrasted and parasternal views (PLAX:LV, PLAX:ID, PSAX:MV), for which the number of training samples is one or two orders of magnitude smaller compared to the apical classes. This showcases the benefit of pretraining for increased performance even with reduced training samples size.

\paragraph{Out-of-distribution testing:} We additionally test the models on two unseen test datasets which contain four and two chamber images. The EchoApex models show better generalization capabilities, with the best performing models being EchoApex-AdaS and EchoApex-B. Detailed results for this experiment are available in the Supplementary Table. \ref{table:zero_shot_view_cls_results}. These results emphasize the generalization capabilities of the EchoApex models.

\subsubsection{View classification with parameter efficient adaptation}
We investigate the use of Adapters \cite{Houlsby2019ParameterEfficientTL} for fine-tuning our EchoApex backbone. Adapters are modules inserted between the frozen backbone's transformer blocks and are useful in reducing the computational cost of fine-tuning a large model, as only the adapter layers are trained. Additionally, they allow for more modularity, as one can fine-tune adapter modules for different tasks while keeping the same backbone.

We report the performance of EchoApex modules with adapters in Fig. \ref{fig:view_cls_gridplot}. Notably, EchoApex-AdaB outperforms not only its counterpart with a Vit-S backbone but also the fine-tuned EchoApex-S ($P < 0.01$ for both cases). In terms of balanced accuracy, EchoApex-AdaB significantly surpasses EchoApex-AdaS in 11 classes (no significant difference in 4 classes) and EchoApex-S in 8 classes (no significant difference in 6 classes). EchoApex-AdaB showing improved performance compared to the fine-tuned EchoApex-S showcases the utility of adapters in achieving a strong performance at a fraction of the computational cost compared to fine-tuning the entire backbone. Tables listing the models' performance and metrics per class for each model are available in Supplementary Tables.
\ref{tab:supp_view_cls_resnet_detailed_results}, \ref{tab:supp_view_cls_vitS_detailed_results}, \ref{tab:supp_view_cls_vitB_detailed_results}.

\subsection{Interactive Structure  Segmentation}\label{sec4:interactivesegmentation}
We sought to evaluate EchoApex's effectiveness in structure segmentation, a common yet time-consuming task in echo exams. We use three large public echocardiogram segmentation datasets, CAMUS \cite{Leclerc2019DeepLF}, EchoNet-Dynamic (ENDym) \cite{Ouyang2020VideobasedAF},  EchoNet-Pediatric (ENPed) \cite{Reddy2023VideoBasedDL}, as well as an internal dataset acquired with volume transducers. CAMUS includes left ventricle and left atrium tracings over an entire cardiac cycle. EchoNet-Dynamic provides tracings of the left ventricle in the apical-4-chamber (A4C) view at end-systole and end-diastole. EchoNet-Pediatric offers tracings of the left ventricle in both A4C and parasternal short axis (PSAX) views. The internal dataset comprises tracings of all four chambers, with biplanes utilized for training and evaluation. The number of annotations per dataset is shown in Fig. \ref{fig:dinosam_distribution}. We used the Dice Similarity Coefficient as our primary evaluation metric, consistent with the public benchmark. Additional details regarding the datasets, experimental setup, and other performance metrics are provided in Methods and Supplementary Table \ref{table:interactivesegmentation_specialistmedsam}.

We build an interactive segmentation model EchoApex-SAM-B (and its variation EchoApex-SAM-S) by incorporating the pretrained EchoApex encoder and prompt-based (including points, box, and text) encoder-decoder modules following SAM \cite{kirillov2023segment}, the state-of-the-art foundation model on natural image segmentation. The model architecture is illustrated in Fig. \ref{fig:dinosam_diagram}. We compare EchoApex-SAM with two types of model. The first is MedSAM \cite{ma2024segment}, a foundation model trained on 1.5 million annotated images from diverse modalities, including CT, MRI, X-ray and ultrasound, and it is specialized in general medical image segmentation. The second are sub-task specialist models, e.g. a UNet model \cite{Leclerc2019DeepLF}, that are individually trained on and published together with each dataset. Results are illustrated in Fig. \ref{fig:dinosam_unet_medsam}. We observe that the UNet specialist models generally shows better performance than MedSAM. This hints that adding additional data from different modalities does not necessarily improve the performance of the model on echo segmentation. As echo represents only a small portion (approximately 5\%) of the training data in MedSAM, its model performance on echo may be impacted negatively by the imbalanced data distribution. We also observe that EchoApex-SAM-B outperforms both MedSAM and UNet-specialist on all categories in all datasets, achieving an average dice score of 0.927 (95\% CI 0.926-0.928), showing an average of 0.23 and 0.57 dice improvement accordingly. This supports the observation that the model pretrained with more in-domain data will boost the performance on the target task.

\subsubsection{Effectiveness of pretraining on few-shot segmentation learning}
It has been shown in prior art that pretraining with large in-domain data yields better performance on few-shot classification. We investigate whether this observation holds for few-shot segmentation learning. We randomly sample 50, 100, 500, 1000 and 5000 annotated images from the whole dataset, and evaluate the performance of EchoApex-SAM-S with and without pretraining on the same evaluation dataset used in the aforementioned public benchmark. Each experiment is repeated 24 times, maximizing the usage of 3 node of 8-GPUs on the compute cluster. From the results in \ref{fig:echoapex_dinosam_fewshot}, we observe that EchoApex does not show superior performance to the imagenet pretrained model when samples are very few, e.g. 50 or 100. However, the performance of pretrained model becomes consistently superior than the imagenet pretrained model when sample size increases more than 500, with an average of 0.042, 0.039, 0.052 dice increment, reaching an average score of 0.893 (500 samples), 0.906 (1000 samples), and 0.923 (5000 samples) respectively. Further, the standard deviation of the EchoApex model (0.060 for 500, 0.053 for 1000 and 0.037 for 5000) is also consistently smaller than the imagenet pretrained model (0.064 for 500, 0.054 for 1000 and 0.052 for 5000), indicating that the model pretrained on in-domain is more stable and robust when the scale of annotations increases.

\subsubsection{Segmentation performance generalizability}
We also assess the generalizability of the segmentation performance, i.e. how well it performs on unseen data. We retrain EchoApex-SAM from scratch on the EchoNet-Dynamic dataset, and compare its performance with the specialist model DeepLabV3 published together with the dataset. We used the ViT-S model to ensure a comparable model size with DeepLabV3. Following Fig. \ref{fig:echoapex_deeplabv3_gen_test}, We observe that EchoApex-SAM shows improved performance compared to DeepLabV3, not only on the in-domain test set from EchoNet-Dynamic, but also on the generalization test set. Specifically, EchoApex-SAM-S achieves an average dice score of 0.923 (95\% CI 0.922-0.925)  on the in-domain test set, and an average dice score 0.877 (95\% CI 0.873-0.880) on the generalization test set, with 0.756, 0.905, 0.889 on Vol-Biplane, CAMUS and EchoNet-Pediatric respectively. The DeepLabV3 model achieves an average dice score of 0.915 (95\% CI 0.913-0.917)  on the in-domain test set, and an average dice score 0.834 (95\% CI 0.829-0.840) on the generalization test set, with 0.744, 0.848, 0.844 on Vol-Biplane, CAMUS and EchoNet-Pediatric respectively. This indicates that the model pretrained on in-domain data, even after fine-tuned, can still be more robust and generalizable than the specialist models.

\begin{figure*}
    \centering
     \hspace{0.2in}
    \begin{subfigure}{0.45\textwidth}
         \centering
         \includegraphics[width=\textwidth]{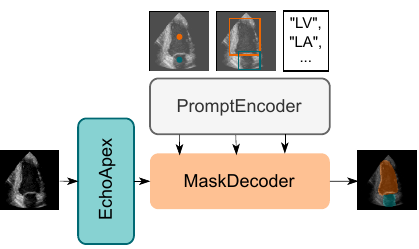}
         \vspace{0.005in}
         \captionlistentry{}
         \label{fig:dinosam_diagram}
     \end{subfigure}
     \PlaceText{30mm}{31mm}{\textbf{(a)}}
     \hfil
    \begin{subfigure}{0.40\textwidth}
         \centering
         \includegraphics[width=\textwidth]{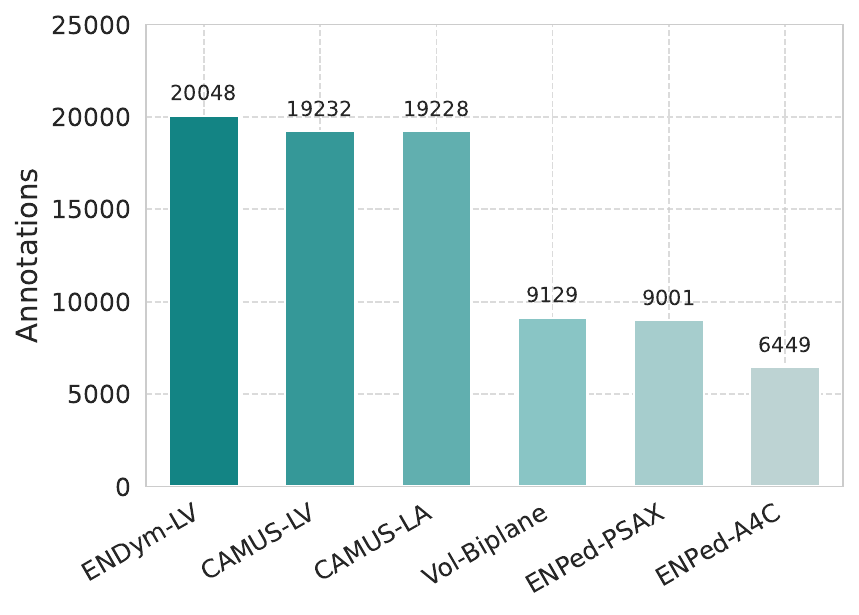}
         \captionlistentry{}
         \label{fig:dinosam_distribution}
     \end{subfigure}
     \PlaceText{117mm}{31mm}{\textbf{(b)}}
    \begin{subfigure}{\textwidth}
         \centering
         \includegraphics[width=0.95\textwidth]{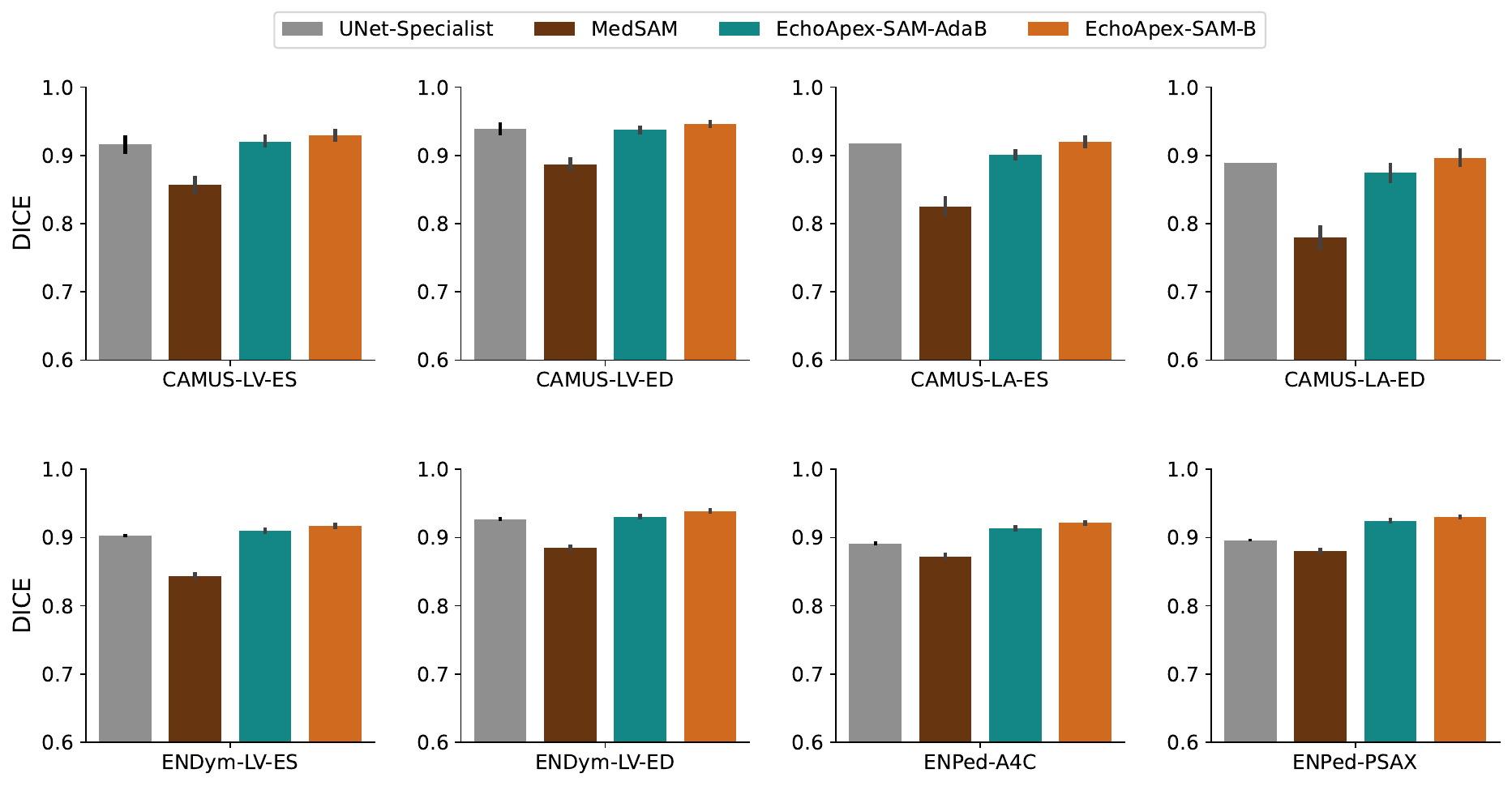}
         \captionlistentry{}
         \label{fig:dinosam_unet_medsam}
     \end{subfigure}
     \PlaceText{30mm}{78mm}{\textbf{(c)}}
    \begin{subfigure}{0.45\textwidth}
         \centering
         \includegraphics[width=\textwidth]{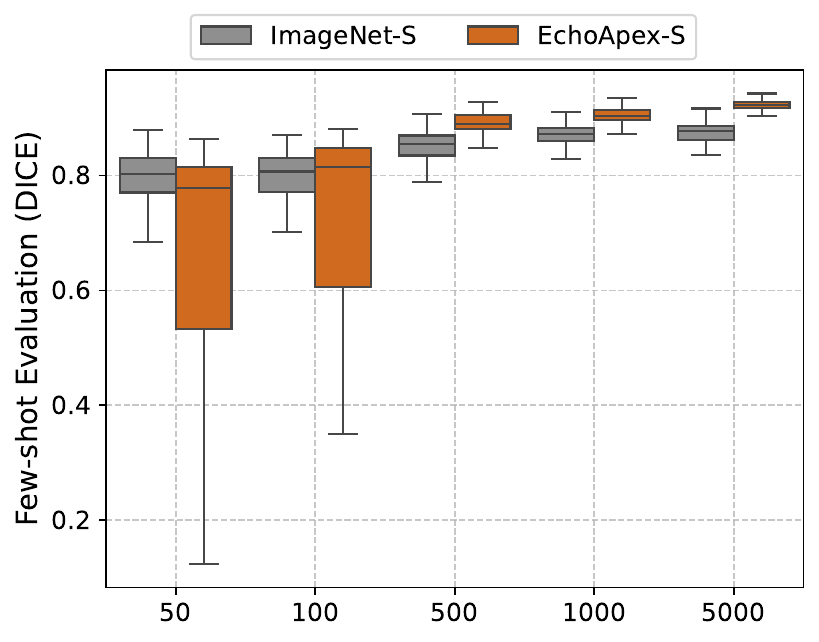}
         \captionlistentry{}
         \label{fig:echoapex_dinosam_fewshot}
     \end{subfigure}
     \PlaceText{30mm}{169mm}{\textbf{(d)}}
     \hfil
    \begin{subfigure}{0.45\textwidth}
         \centering
         \includegraphics[width=\textwidth]{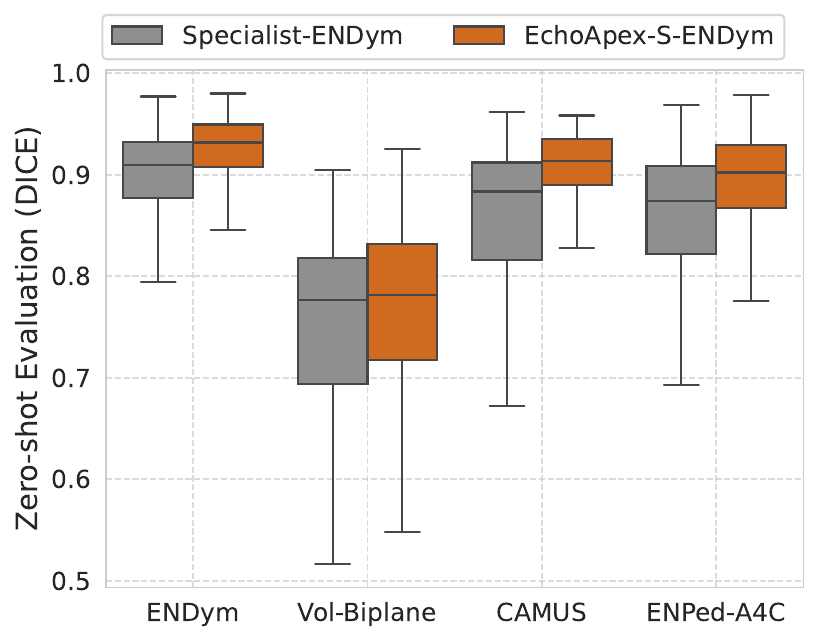}
         \captionlistentry{}
         \label{fig:echoapex_deeplabv3_gen_test}
     \end{subfigure}
     \PlaceText{110mm}{169mm}{\textbf{(e)}}
     \caption{\textbf{Study on interactive segmentation.} \textbf{(a)} EchoApex attaches a prompt encoder and a mask decoder for the interactive segmentation task, taking prompt in forms of points, boxes and texts. \textbf{(b)} Number of annotations in all evaluated dataset. \textbf{(c)} Performance comparison between EchoApex-SAM variations and sub-task specialist models, e.g. UNet individually trained on each dataset, and generalist model MedSAM trained on multi-modal medical images. Oracle box prompt is used in this experiment category. \textbf{(d)} Few-shot learning evaluation of EchoApex-S vs. its ImageNet pretrained counterpart. Text prompt is used in this experiment category. \textbf{(e)} Generalization capability test of EchoApex-S compared with specialist model DeepLabV3 on both in-domain (ENDym) and out-domain data.}
\end{figure*}

\subsection{Landmark Detection for Left Ventricle Measurements}\label{sec5:lvh}
Left ventricular hypertrophy (LVH) is an adaptation of the cardiac muscle in response to disease, cardiac wall stress or significant hemodynamic pressure. Its diagnosis is crucial as patients with LVH are at increased risk of developing heart failure, arrhythmia, strokes and sudden death \cite{Drazner2011ThePO}. Left ventricular wall measurements using echocardiography in the parasternal long-axis (PLAX) view are widely used in the diagnosis of LVH. However these measurements are subject to inter-observer and intra-observer variability due to the complexity of the myocardial shape, imaging artefacts and image quality \cite{Phelan2017ComparisonOV}.

In clinical practice, the LV mass is indexed against the body surface area to and compared to reference values to perform a diagnosis of LVH and determine its severity \cite{Barbieri2012LeftVH}. Three key measurements are taken in the PLAX view to compute the LV mass: The intraventricular septum (IVS), the LV internal dimension (LVID) and the LV posterior wall thickness (LVPW). AI-driven approaches in medical imaging have demonstrated significant potential in enhancing the accuracy and consistency of cardiac measurements. AI models can analyze echocardiographic images with a high degree of precision, minimizing human error and subjectivity and thus benefiting the clinical workflows \cite{Howard2021AutomatedLV,Augusto2020DiagnosisAR}.

\begin{figure*}
    \centering
    \begin{subfigure}{0.47\textwidth}
         \centering
         \includegraphics[width=\textwidth]{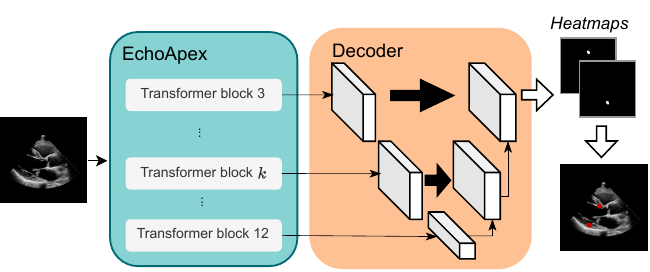}
         \vspace{0.01in}
         \caption{}
         \label{fig:lvh_diagram}
     \end{subfigure}
     \hfil
         \begin{subfigure}{0.38\textwidth}
         \centering
         \includegraphics[width=\textwidth]{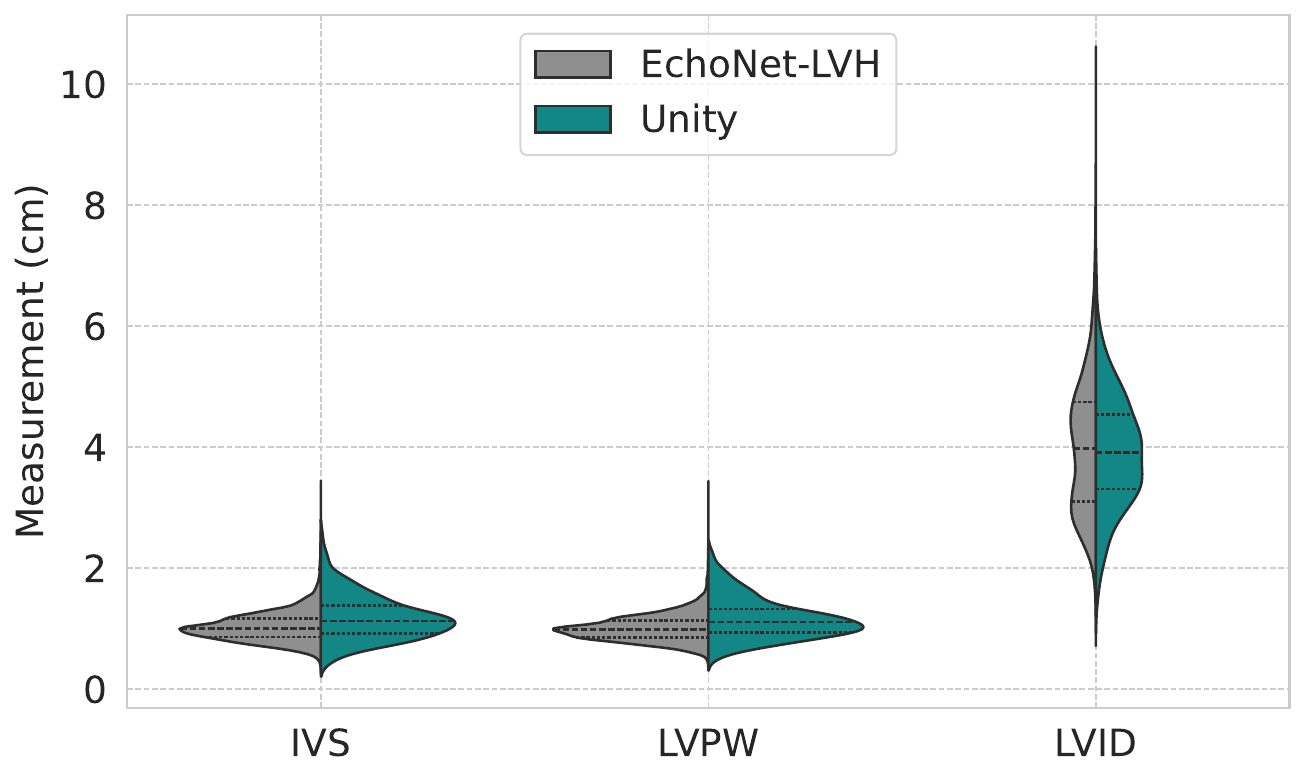}
         \caption{}
         \label{fig:lvh_dset_distribution}
     \end{subfigure}
     \hfill

    \begin{subfigure}[b]{0.45\textwidth}
         \centering
         \includegraphics[width=\textwidth]{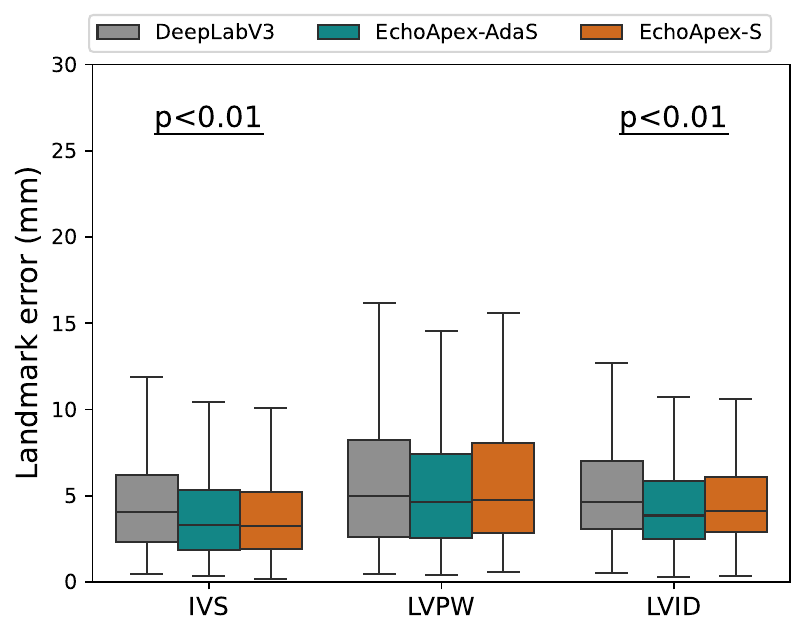}
         \caption{EchoNet-LVH}
         \label{fig:lvh_lmk_box_results}
     \end{subfigure}
     \begin{subfigure}[b]{0.45\textwidth}
         \centering
         \includegraphics[width=\textwidth]{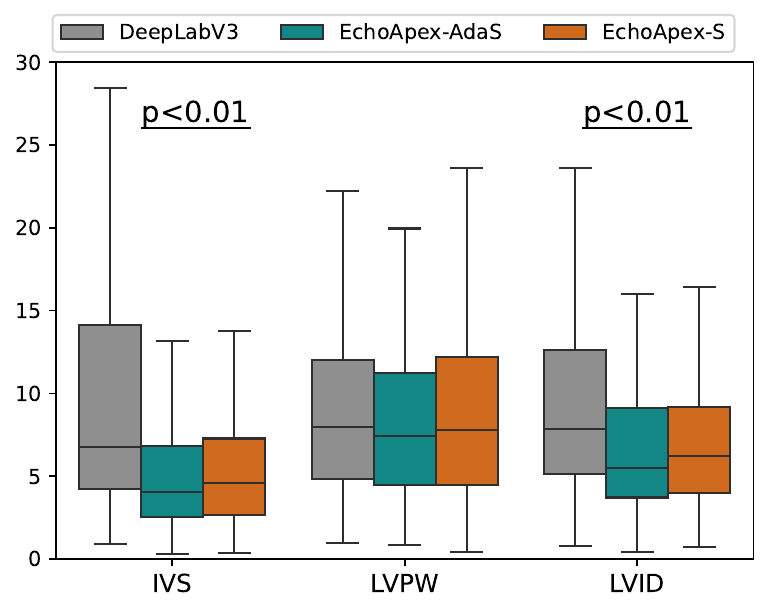}
         \caption{Unity (Out-of-distribution)}
         \label{fig:unity_lmk_box_results}
     \end{subfigure}

    \begin{subfigure}[b]{0.45\textwidth}
         \centering
         \includegraphics[width=\textwidth]{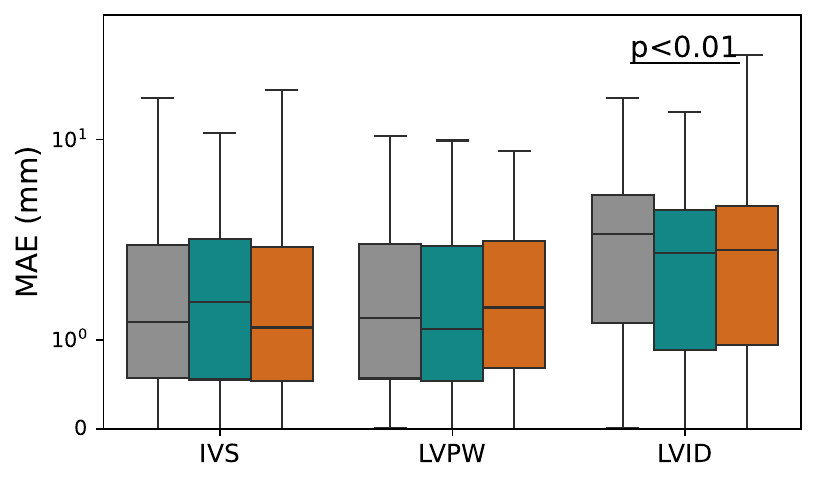}
         \caption{EchoNet-LVH}
         \label{fig:lvh_mae_box_results}
     \end{subfigure}
     \begin{subfigure}[b]{0.45\textwidth}
         \centering
         \includegraphics[width=\textwidth]{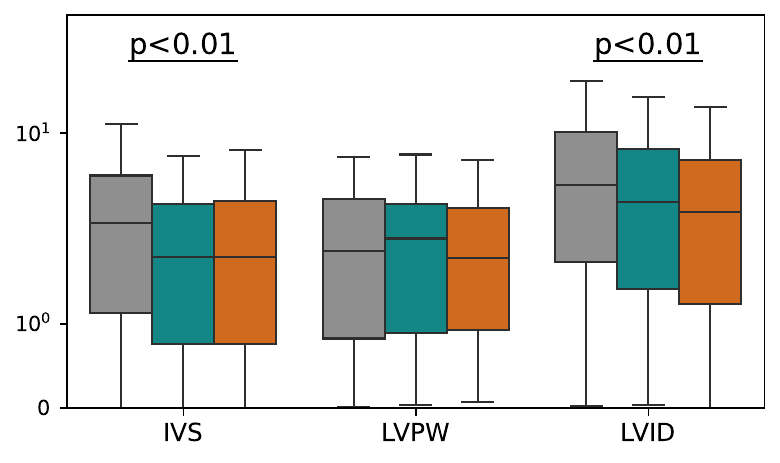}
         \caption{Unity (Out-of-distribution)}
         \label{fig:unity_mae_box_results}
     \end{subfigure}
     
    \begin{subfigure}[b]{0.8\textwidth}
         \centering
         \includegraphics[width=\textwidth]{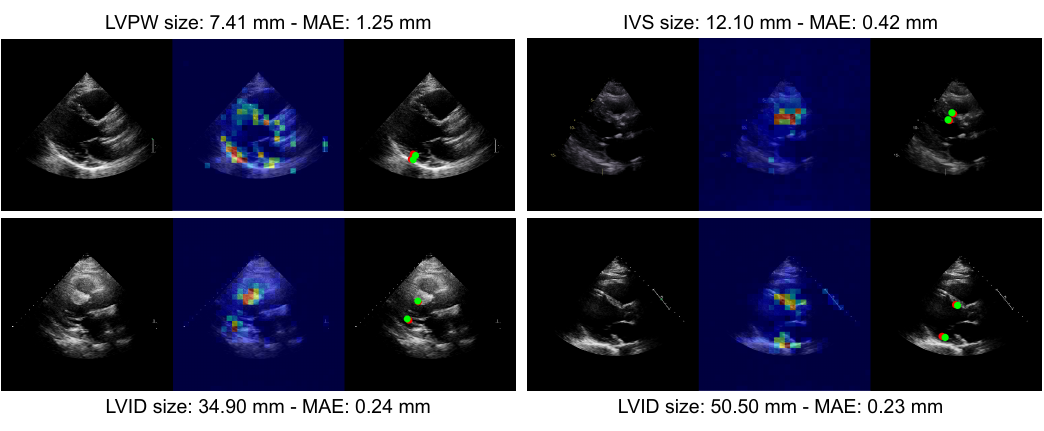}
         \vspace{-0.4in}
         \caption{}
         \label{fig:lvh_example_attention}
     \end{subfigure}
    
    \caption{\textbf{Study on left ventricle hypertrophy detection.} \textbf{(a)} Architecture for the landmark detection task. \textbf{(b)} Measurements distribution for the EchoNet-LVH (internal) and Unity (external) datasets. \textbf{(c, d)} Landmark error (mm) and \textbf{(e, f)} MAE (mm) distribution on the test datasets. $p<0.01$ is from a one-sided t-test showing significant improvement between  EchoApex-S and DeepLabV3. \textbf{(g)} Self-attention maps from the last block of the EchoApex encoder showing regions of high similarity between image tokens, alongside the predicted (red) and ground truth (green) landmark positions.}
    \label{fig:lvh_results}

\end{figure*}

\subsubsection{Supervised landmark detection for left ventricle wall thickness measurements}

We use our pre-trained EchoApex model as an image encoder and a decoder architecture inspired by UNETR \cite{hatamizadeh2022unetr}, as described in Fig. \ref{fig:lvh_diagram}. We compare our model with the state-of-the-art method from \cite{Duffy2021HighThroughputPP} which uses a DeepLabV3 \cite{Chen2017RethinkingAC} backbone, retrained following the authors' specifications. All hyperparameters are listed in the Supplementary Table. \ref{table:supp_lvh_hyperparams}. Training is conducted on the training split of the EchoNet-LVH dataset \cite{Duffy2021HighThroughputPP} while testing is performed on two datasets: the test split of EchoNet-LVH (internal dataset) and the validation split of the Unity dataset \cite{Howard2021AutomatedLV} (external dataset). Fig. \ref{fig:lvh_dset_distribution} details the distribution of the measurements for the two datasets.

We report the average landmark error (L2 norm) between all landmarks and the Mean Absolute Error (MAE) on the three key measurements on the internal EchoNet-LVH (resp. Fig. \ref{fig:lvh_lmk_box_results} and Fig. \ref{fig:lvh_mae_box_results}) and external Unity test datasets (resp. Fig. \ref{fig:unity_lmk_box_results} and Fig. \ref{fig:unity_mae_box_results}). Qualitative results showing attention maps and landmark predictions are shown in Fig. \ref{fig:lvh_example_attention} and in the Supplementary Fig. \ref{fig:supp_lvh_qualitative_results}.

EchoApex-S achieves superior performance compared to the DeepLabV3 baseline on the internal EchoNet-LVH dataset. Landmark error is significantly improved on IVS and LVID measurements, with $P < 0.01$, according to a one-sided t-test. We report mean values of 4.11 mm and 4.89 mm for IVS and LVID, respectively, for our model, and DeepLabV3 achieves values 4.69 mm and 5.35 mm, respectively. MAE on LVID measurements is also reduced for EchoApex-S with a value of 2.71 mm compared to a value of 3.23 mm for DeepLabV3. We report $R^2$ values between the model and human measurements of 0.96 and 0.95 for the EchoApex-S and DeepLabV3 models, respectively.

On the external Unity dataset, EchoApex-S achieves substantial improvement on the IVS and LVID measurements compared to DeepLabV3 ($P$ value $<$ 0.01), both for landmark error and MAE, with a mean landmark errors of 5.98 mm and 7.57 mm on IVS and LVID, as opposed to 13.86 mm and 11.25 for the DeepLabV3 baseline. MAE values obtained by EchoApex-S are 2.69 mm and 4.82 mm for IVS and LVID, respectively while the DeepLabV3 performance on the unseen test distribution is worse, with MAE values of 3.78 mm and 7.54 mm for IVS and LVID, respectively. EchoApex-S achieves a $R^2$ of 0.87 on the external test dataset, as opposed to 0.72 for DeepLabV3. Detailed results are available in the Supplementary Tables. \ref{table:supp_lvh_mae_results}-\ref{table:supp_unity_lmk_results}. These findings indicate that pretraining on the extensive Echo20M dataset enhances generalization to unseen data distributions, as demonstrated by the results on the external Unity dataset. Furthermore, the results on EchoNet-LVH highlight that large-scale pretraining improves both robustness and accuracy.

\subsubsection{Left ventricle measurements with parameter efficient adaptation}

We investigate the use of adapters for parameter efficient tuning of the EchoApex backbone. As previously done, we freeze the EchoApex encoder and finetune only the adapter layers. On the internal EchoNet-LVH dataset, EchoApex-AdaS outperforms significantly ($P < 0.01$) the DeepLabV3 baseline in both landmark error and MAE for the IVS and LVID measurements. We report at $R^2$ of 0.81 for EchoApex-AdaS on this dataset.

On the external Unity dataset, EchoApex-AdaS shows superior performance compared to DeepLabV3, with significantly better results for both landmark error and MAE in the measurements of IVS and LVID ($P < 0.01$) for all). There are no statistically significant difference between the performance of the EchoApex-S and EchoApex-AdaS on both datasets except for the MAE on the Unity LVID measurements, where EchoApex-AdaS outperforms its fine-tuned counterpart. However. EchoApex-S shows a better agreement in terms of measurements compared to human annotators with a $R^2$ of 0.87 on the Unity dataset, as opposed to 0.81 for its adapter counterpart.

EchoApex-AdaS comparable performance to EchoApex-S on the EchoNet-LVH dataset shows the utility of the parameter efficient tuning in resource limited settings. In this experiment, the training dataset was orders of magnitude smaller compared to the view classification task. Indeed, tens of thousand of training samples were used for this task as opposed to millions in the view classification experiment. However, the EchoApex-S predictions agree better with human annotations on the Unity dataset, indicating a less effective generalization of EchoApex-AdaS to the unseen distribution.

    \subsection{Automatic Ejection Fraction Estimation}\label{sec6:autoef}
\begin{figure*}
    \centering
    \begin{subfigure}{0.45\textwidth}
         \centering
         \includegraphics[width=\textwidth]{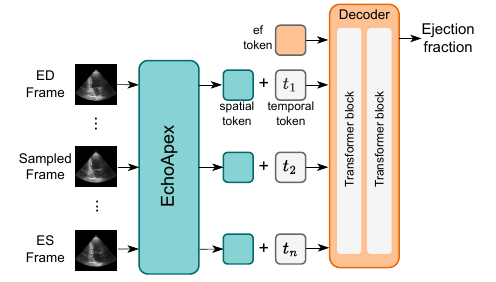}
         \caption{}
         \label{fig:autoef_diagram}
     \end{subfigure}
         \begin{subfigure}{0.4\textwidth}
         \centering
         \includegraphics[width=0.9\textwidth]{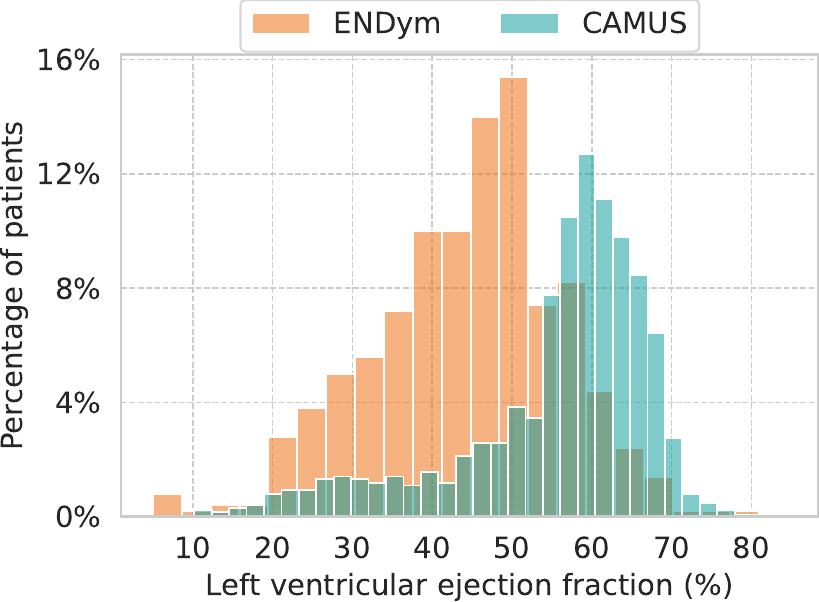}
         \caption{}
         \label{fig:autoef_ef_distribution}
     \end{subfigure}
     \hfill

    \begin{subfigure}[b]{0.4\textwidth}
         \centering
         \includegraphics[width=\textwidth]{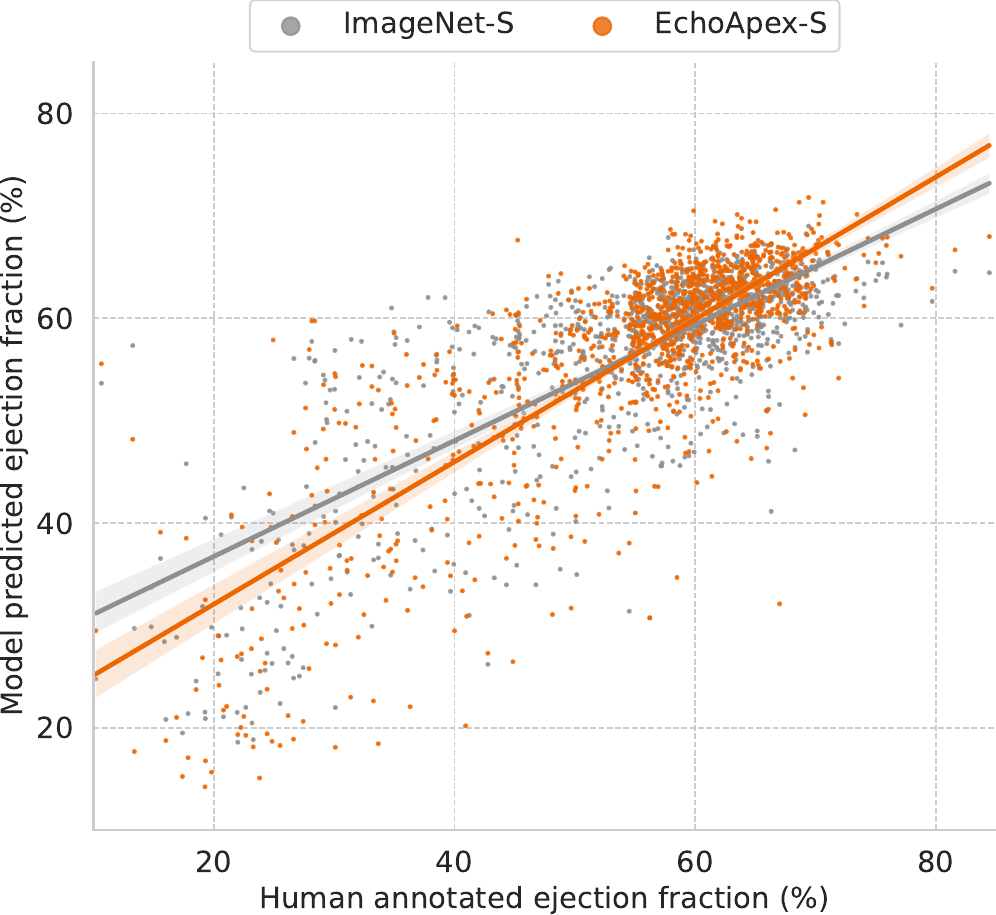}
         \caption{EchoNet-Dynamic}
         \label{fig:autoef_echonet_scatter}
     \end{subfigure}
     \hfil
     \begin{subfigure}[b]{0.4\textwidth}
         \centering
         \includegraphics[width=\textwidth]{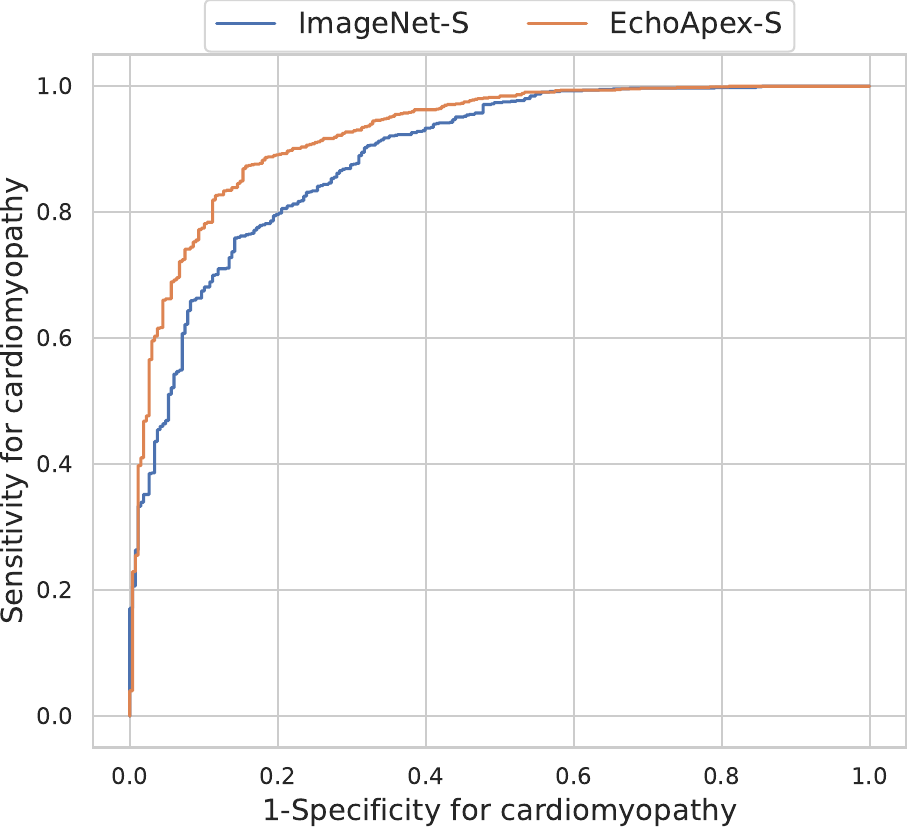}
         \caption{EchoNet-Dynamic}
         \label{fig:autoef_echonet_roc}
     \end{subfigure}
        \begin{subfigure}[b]{0.4\textwidth}
         \centering
         \includegraphics[width=\textwidth]{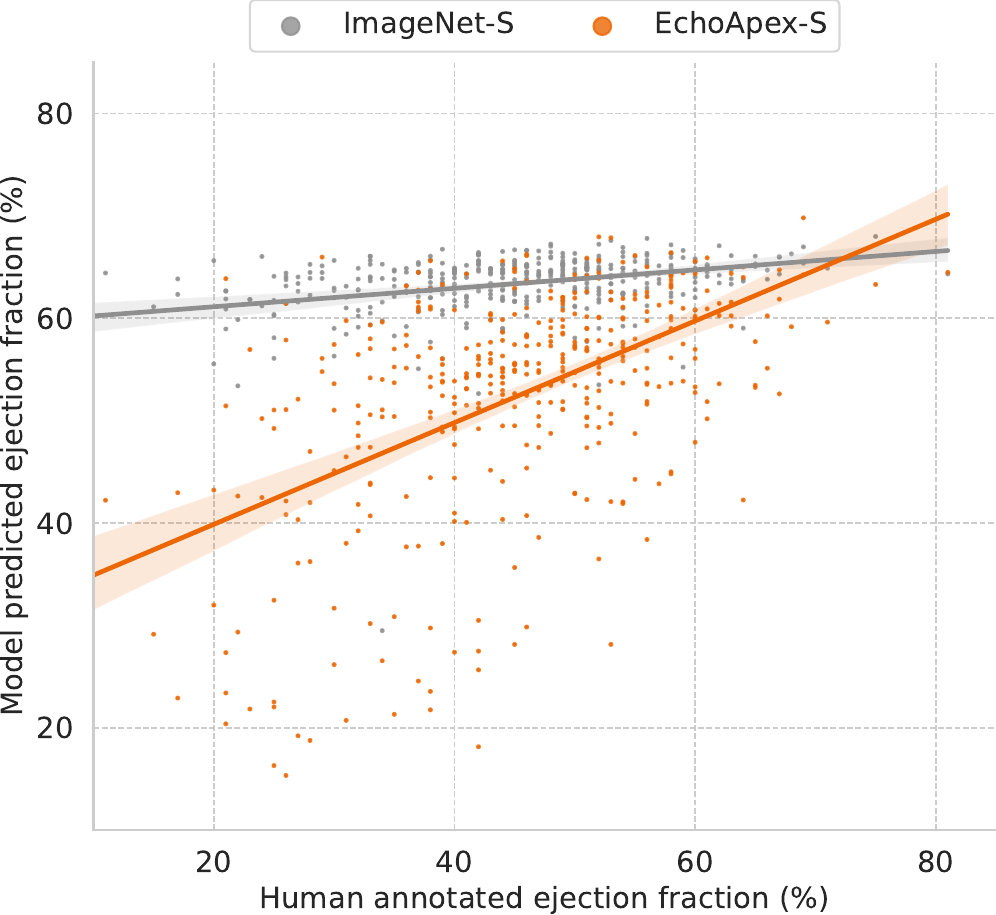}
         \caption{CAMUS (zero-shot)}
         \label{fig:autoef_camus_scatter}
     \end{subfigure}
     \hfil
     \begin{subfigure}[b]{0.4\textwidth}
         \centering
         \includegraphics[width=\textwidth]{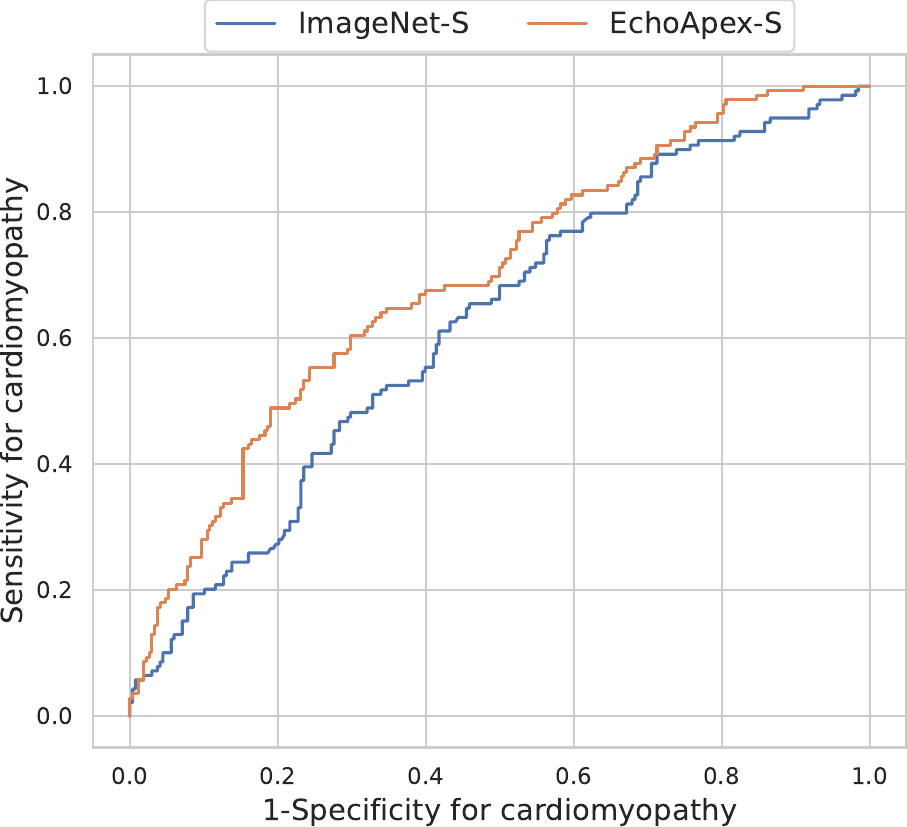}
         \caption{CAMUS (zero-shot)}
         \label{fig:autoef_camus_roc}
     \end{subfigure}
    \caption{\textbf{Study on automated ejection fraction estimation.} \textbf{(a)} EchoApex model extension for spatial-temporal prediction EchoApex-ST. The model takes a heart beat as input and directly predicts the ejection fraction value. \textbf{(b)} Distribution of LVEF for the EchoNet-Dynamic (internal) and CAMUS (external) datasets. \textbf{(c, d)} Performance comparison between EchoApex-S and its ImageNet pretrained counterpart on EchoNet-Dynamic dataset. Both models have been finetuned on this datase. (c) shows the predicted EF value versus human annotated ground truth. (d) shows ROC of cardiomyopathy using a threshold of 50\%.  \textbf{(e, f)} Same evaluation of both models on CAMUS (unseen) dataset.}
    \label{fig:autoef_results}

\end{figure*}

We assessed the performance of the proposed model in the task of automated ejection fraction (EF) estimation. EF, defined by changes in blood volume, inherently requires a model with a temporal perspective. Although the model was initially pretrained on individual images, we aimed to evaluate the potential of EchoApex for temporal prediction by extending it with a spatial-temporal architecture, termed as EchoApex-ST and illustrated in Fig. \ref{fig:autoef_diagram}. We train EchoApex-ST on the public EchoNet-Dynamic dataset and test it on both the same dataset for in-domain evaluation and CAMUS dataset for out-domain evaluation to assess its generalization capability. We compare EchoApex-ST with its ImageNet pretrained counterpart ImageNet-ST. Throughout the finetuning process, we froze the image encoder of both models and only trained the temporal decoder. 

On the EchoNet-Dynamic dataset, EchoApex-ST achieved a mean absolute error (MAE) of 5.6\% for LVEF, reducing the error by 0.5\% compared to ImageNet-ST (6.1\%). When using an LVEF threshold of less than 50\% to classify cardiomyopathy, EchoApex-ST had an area under the curve (AUC) of 0.93, whereas ImageNet-ST had an AUC of 0.89(Fig. \ref{fig:autoef_echonet_scatter}, \ref{fig:autoef_echonet_roc}). On the CAMUS dataset, where neither EchoApex-ST nor ImageNet-ST were trained, EchoApex-ST achieved an MAE of 10.7\% for LVEF, a reduction of 7.7\% compared to ImageNet-ST's 18.4\%. Using the same LVEF threshold of less than 50\% for cardiomyopathy classification, EchoApex-ST again had an AUC of 0.69, while ImageNet-ST had an AUC of 0.62 (Fig. \ref{fig:autoef_camus_scatter}, \ref{fig:autoef_camus_roc}).   

\section{Discussion}\label{sec7:discussion}

Modern deep learning models have significantly advanced the study and clinical applications of echocardiography. Our study on EchoApex demonstrates that a general-purpose vision foundation model, when pretrained on a large and diverse dataset, can outperform specialized task-specific deep learning models across a range of clinical applications. The comprehensive evaluation of EchoApex across four distinct clinical applications underscores its versatility and robustness, highlighting the value of foundation models in medical imaging.

EchoApex’s impressive performance across various tasks emphasizes the potential of large-scale pretraining with in-domain data. To ensure a diverse range of data for pretraining, we curated Echo20M, a dataset containing 20 million images derived from 450,338 videos across 11 clinical centers. By varying data and model scales, we observed a consistent pattern in echocardiography similar to that in natural images and other medical imaging domains: downstream task performance improves as data and model size increase. In a K-Nearest Neighbors (KNN) test classification, the EchoApex model achieved an accuracy of 0.92 without any label supervision during training. When trained with labeled data, the balanced accuracy (BACC) improved to 0.98, surpassing models pretrained on ImageNet in 14 out of 18 views. This result highlights the effectiveness of leveraging large-scale in-domain data for specific downstream tasks with limited supervision.

EchoApex also excelled in segmentation tasks when trained on a large annotated dataset, achieving a mean score of 0.93. This outperformed specialist models and the generalist model MedSAM, which scored 0.90 and 0.88, respectively. These results validate the model's ability to utilize supervised learning effectively. In few-shot evaluation, EchoApex demonstrated stable and superior performance compared to its ImageNet-pretrained counterparts. In zero-shot evaluation, EchoApex not only surpassed specialist models on the trained in-domain dataset but also on external datasets, demonstrating superior generalizability even after fine-tuning. This observation also applies to left ventricle measurement tasks, where EchoApex achieved better in-domain performance in both landmark detection and ventricle measurement, with an average MAE of 0.2 mm compared to specialist baselines. Notably, it showed a remarkable improvement of 1.4 mm on 1K frames from an unseen data cohort. Further, EchoApex's success in achieving high accuracy and AUC scores in tasks such as ejection fraction estimation and cardiomyopathy detection, even with a frozen encoder, emphasizes its ability to generalize across different datasets and tasks without extensive retraining. This generalizability is crucial for real-world clinical applications, where models must handle a wide variety of image types and patient conditions.

While fully fine-tuning a large model end-to-end can be computationally expensive, we explored parameter-efficient adaptation of EchoApex. With less than 4\% of trainable parameters and only a portion of the computational load, the adapted model still showed competitive performance, with a maximum degradation of 5.1\% compared to fully fine-tuned models. This finding suggests the potential to develop an affordable AI system where a unified image encoder is utilized with task-dependent adapters requiring minimal parameters and computation, thereby enhancing performance across a variety of downstream tasks.

Our study has several limitations. EchoApex is pretrained on static images independently, while echocardiography naturally involves sequences of images. This approach may lack a temporal perspective, as shown in the experiment on ejection fraction estimation. Although a sequence decoder was attached to aggregate temporal information in late feature fusion, EchoApex achieved an AUC of 0.93 in cardiomyopathy classification. While this is higher than the ImageNet-pretrained model's AUC of 0.89, it is lower than the specialized spatiotemporal model with early feature fusion, which achieved an AUC of 0.97 \cite{Ouyang2020VideobasedAF}. Across all tasks, we used the same adapters. While the adapted model already showed competitive performance with task-specific models, having task-specific adapters such as ViT-Adapter \cite{chen2022vision} and Adaptformer \cite{chen2022adaptformer} could potentially lead to further improvement based on the characteristics of the downstream task. Besides model development, fairness and equity considerations are essential. For privacy reasons, much of the evaluation dataset, including both internal and external ones, had sex and demographic information anonymized. This limits our ability to evaluate the impact of the pretraining distribution on downstream tasks.

The promising results of EchoApex suggest several potential avenues for future research. One key direction is to explore the integration of multimodal data \cite{singhal2023large, lu2024visual,christensen2024vision}, such as combining echocardiographic images with other clinical data, including cardiac CT and clinical reports. This could further enhance model performance and provide more comprehensive diagnostic insights. Another important avenue is to distill the large model into more efficient versions capable of performing real-time inferences for procedural guidance in structural heart interventions and electrophysiology procedures. Lastly, ensuring the model's fairness and generalizability across diverse populations is crucial \cite{jin2024fairmedfm}. Future research should examine the influence of demographic variables and disease profiles on model performance to ensure equitable outcomes.

\clearpage

\begin{appendices}

\section{Methods}\label{secA1}

The following sections are organized as follows: for each core component of EchoApex, including pretraining, view classification, structure segmentation, ventricular measurement, and automated ejection fraction estimation, we present the details of the data curation process, model architecture, training, and evaluation.

\subsection{Pretraining}

\textbf{Curation of EchoApex dataset}. To ensure the quality and diversity of data for model pretraining, we curated a large-scale dataset of echocardiographic images from 450,338 videos involving 26,704 patients across 11 clinical centers. This dataset encompasses a comprehensive range of echocardiographic data, including transthoracic echocardiograms (TTE) from routine diagnostic exams, as well as transesophageal echocardiograms (TEE) and intracardiac echocardiograms (ICE) obtained from various echo-guided cardiac interventions. Data were acquired using transducers from multiple vendors: Siemens Healthineers (68\%, by number of videos), GE (15\%), Philips (16\%), and Samsung (1\%). All images underwent a de-identification process to remove any protected health information (PHI) and were stored in DICOM format. We developed a preprocessing workflow to remove non-image data, including ECG signals, text, and respirometer information. For DICOM files acquired with volume transducers, we converted them into Cartesian space and then extracted 2D slices from the volume data. For each volume, 12 slices were extracted at 15-degree intervals from the AB-plane views (sagittal and coronal views), and 3 slices from the C-plane view (transverse view). For both the 2D echo and sliced volume images, we performed a tight crop to the image region, padded to square, and resized to $112\times 112$ pixels. A total of 37.4 million images were extracted from the entire dataset. As described in the main body of the paper, to study the impact of dataset size on model performance, we sampled the dataset into Echo3M, Echo12M, and Echo20M. Although the dataset for the downstream tasks could come from the same transducers of the same vendors as the pretraining dataset, we deliberately avoided any overlap between the patients in the pretraining and all downstream evaluation sets to minimize the risk of data contamination.

\textbf{Protocol of EchoApex pretraining}. We used the Vision Transformer (ViT) architecture \cite{dosovitskiy2020image} for pretraining. In terms of model size, we utilized the base model, ViT-B, which has 12 blocks, a patch size of 14, and a feature embedding dimension of 768, as well as the smaller model, ViT-S, which has 12 blocks, a patch size of 8, and an embedding dimension of 384. We leveraged the state-of-the-art self-supervised learning approaches from the DINO method family (DINOv1 \cite{caron2021emerging} and DINOv2 \cite{oquab2023dinov2}) for model pretraining. The DINO methods are based on teacher-student self-distillation, where two augmented images from the same input are processed through the teacher and student networks, respectively, and the student network is trained to predict the teacher network's output. Compared to DINOv1, DINOv2 integrates additional best practices, such as iBOT loss \cite{zhou2021ibot}, Sinkhorn-Knopp centering \cite{caron2020unsupervised}, KoLeo regularization \cite{sablayrolles2019spreading}, and computational optimizations. While these components in DINOv2 are designed to improve performance, they also increase computational cost. To balance computation and convergence, we employed a two-stage approach: starting with DINOv1 until the pretrained model converged and then continuing training with the full DINOv2. Additionally, to leverage knowledge from large-scale natural images, we initialized our pretraining model with pretrained weights from the officially released DINO checkpoints. Since DINOv2 does not release a pretrained checkpoint for ViT-S with a patch size of 8, we used the pretrained ViT-S model from DINOv1 instead. We followed the same training hyperparameters as detailed in the DINO papers, which are outlined in Table \ref{tab:pretraining_hyperparameters}.  

\subsection{View Classification}

\textbf{Curation of view classification datasets}
The datasets used for view classification come from 6 different clinical sites and include several vendors, including Siemens, GE, Philips and Samsung. All patient data was removed from the DICOM files. All images were extracted from video clips and processed following these steps: Firstly, we extracted the B-mode image from the DICOM, discarding extra information in the image. We masked all the data outside the imaging cone to remove potential annotations, texts and elements from the user interface. The images were then padded to square sizes and finally converted to grayscale. Images were resized to 112x112 for the ViT-B models, 128x128 for the ViT-S models and the ResNet50 baseline. The data was split in the following way for model development (number of video sequences / number of images): Train (21500 / 2.5M) ; Validation (2150 / 250K) ; Test (2100 / 250K).

\textbf{View classification with linear decoder}
View classification is a pivotal task for echocardiographic interpretation and several works have investigated it, mostly focusing on TTE for standard view recognition \cite{Madani2018FastAA,Madani2018DeepED,stvik2019RealTimeSV,Zhang2018FullyAE}. Authors in Gearhart et al. \cite{Gearhart2022AnAV} investigate this task for pediatric patients. Other ultrasound modalities such as transesophageal echocardiography are less explored \cite{Steffner2023DeepLF}.

The decoder employed for this task was a linear layer which took as input the features from [CLS] (that is, classification) token of the last ViT block of the EchoApex encoder. During training, we randomly sampled images from each video sequence to form a training batch. Images were normalized to the [-1,1] range before being inputted in the image encoder. A weighted cross-entropy loss based on class probabilities was used to fight class imbalance. The models selected for the experiments were the ones with the lowest validation loss. The hyperparameters used for this experiment are specified in Table. \ref{table:supp_view_cls_hyperparams}.

When evaluating, for each video sequence, we used 5 frames from the sequence and performed a majority voting to determine the label associated to the sequence. Namely, given a sequence with $N$ frames, we selected the frames at index 0, 0.25$N$, 0.5$N$, 0.75$N$ and $N-1$.

\subsection{Structure segmentation}

\textbf{Curation of segmentation datasets}
The EchoNet-Dynamic dataset \cite{Ouyang2020VideobasedAF}, released by Stanford University School of Medicine, comprises 10,030 apical-4-chamber echocardiography videos from 10,030 patients over a three-year period. Each video includes two traces of the left ventricle at end-systole and end-diastole. The CAMUS dataset \cite{Leclerc2019DeepLF}, released by the University of Lyon, includes 1,000 echocardiography videos from 500 patients. It contains apical-4-chamber and apical-2-chamber views, with each view including two traces of the left ventricle and left atrium at end-systole and end-diastole. The EchoNet-Pediatric dataset \cite{Reddy2023VideoBasedDL}, also released by Stanford University School of Medicine, includes 3,176 apical-4-chamber echocardiography videos and 4,424 parasternal short axis echocardiography videos, with a spanning period of 7 years (2014-2021). Each video includes tracings of left ventricle at end-systole and end-diastole by human experts. For all the public dataset, We used the same training and testing splits as is in the original publication for our experiment setup. The internal dataset comprises 975 volumetric transthoracic echocardiograms (TTE) and 195 volumetric transesophageal echocardiograms (TEE). Three expert sonographers performed 3D mesh tracing of the four chambers, reaching a consensus. For each volume, we extracted a pair of perpendicular biplane views from the volumetric data for training and testing. Detailed information on the datasets is shown in Table \ref{table:interactivesegmentation_datadetails}.

\textbf{Interactive Segmentation with SAM.} We built the structure segmentation model using the SAM framework, a state-of-the-art interactive segmentation model \cite{kirillov2023segment,ravi2024sam2}. User inputs are taken as "prompts," and the model provides segmentation output based on these prompts. Prompts can be a bounding box, a point set, or a text description. The flexibility of this prompt-based framework allows the model to be suitable for segmenting a variable number of structures. Our EchoApex-SAM model consists of three modules: the pretrained EchoApex encoder, a prompt encoder, and a mask decoder. The prompt encoder design is similar to that of SAM, with the addition of one convolutional layer and one LayerNorm layer in the mask embedding to match the output dimension to the EchoApex embedded feature dimension. We used the CLIP model \cite{radford2021learning} for text prompt embedding. During finetuning, the CLIP model was frozen, and the rest of the EchoApex-SAM model was trained. We used a combination of cross-entropy loss and Dice loss with equal weights for the training process. Additionally, we employed a multi-forward loss approach, which iteratively updates the prompts with previous mask predictions and aggregates the losses from each forward pass. We found that a forward factor of 4 and a decayed loss weight of 0.9 yielded the best performance.

\textbf{Fully supervised finetuning, few-shot, and zero-shot evaluation.} We evaluated the performance of EchoApex-SAM across three types of studies. To mitigate performance bias from model size, we compared models of the same size as the benchmarked methods in existing literature. In the fully supervised study, we compared EchoApex-SAM-B with MedSAM \cite{ma2024segment}, both using ViT-B as the image encoder. We trained EchoApex-SAM-B and its adapted version EchoApex-Ada-B on all available training datasets with box prompts. During training, box prompts were generated from ground truth annotations and then randomly perturbed up to 15\% translation at each corner. The input size to EchoApex-SAM-B was the same as during pretraining, $112\times112$ pixels. We used the officially released MedSAM checkpoint and resized the input and output to $112\times112$ for a fair comparison. The results for the UNet specialist models were taken directly from the original publications \cite{Ouyang2020VideobasedAF,Leclerc2019DeepLF,Reddy2023VideoBasedDL}. For the few-shot study, we compared EchoApex-SAM-S with its ImageNet pretrained counterpart model. During dataset sampling, we sampled subsets proportionally to the size of each training dataset to match the data distribution. We used the same testing set, prompt augmentation, and training hyperparameters as in the fully supervised study, employing the ground truth box oracle for evaluation. For the zero-shot study, to mitigate potential bias from box prompts, we trained EchoApex-SAM-S on the same training dataset as the DeepLabV3 model, using text prompts. We fixed the text prompt to "left ventricle" throughout the training process and used the officially released DeepLabV3 checkpoint from EchoNet-Dynamic \cite{Ouyang2020VideobasedAF}. For all training processes, we used the AdamW optimizer with a learning rate of $5e-4$, weight decay of 0.005, and a linear learning rate decay schedule. All models were trained for 60 epochs, and the model with the best validation performance was selected for evaluation.

\subsection{Ventricular measurements}
\textbf{Curation of datasets for left ventricle measurements}
Two datasets were preprocessed for this experiment: EchoNet-LVH \cite{Duffy2021HighThroughputPP}, where we used the official train / validate / test split for model development (with respectively 20254, 2275 and 683 annotated frames) and
the Unity dataset \cite{Howard2021AutomatedLV}, where the validation split was employed to evaluate the model performance on 1095 unseen frames.
For the both datasets, all images were padded to obtain square sizes and then resized to 256x256 and to 266x266 for ViT-S and ViT-B experiments, respectively. When training the DeepLabV3 baseline from \cite{Duffy2021HighThroughputPP}, the images were resized to 640x640 pixels.

While the annotations contain two landmark locations for each measurements, when training EchoApex models, we do not use LVID landmarks, as their position can be deduced from
the LVPW landmark furthest from the posterior wall and from the IVS landmark on the LV side. We found this improved performance compared to using all six landmarks for training.
We used the landmark locations to create ground truth heatmaps by generating gaussian balls with variable standard deviation (parameter $\sigma$ in the Table. \ref{table:supp_lvh_hyperparams}).
Each heatmap corresponded to one channel of the label tensor.

\textbf{Landmark detection for left ventricle measurements}
For this task, EchoApex models were trained using a combination of generalized DICE \cite{Sudre2017GeneralisedDO} and focal losses \cite{Lin2017FocalLF}, where each loss contributed equally.
The DeepLabV3 model was trained with a weighted MSE loss, taking in account the imbalance between the number of background and foreground pixels, as done in \cite{Duffy2021HighThroughputPP}. When training, if a given landmark was not annotated on an image, the channel corresponding to that landmark was masked.
As per the authors' code, we used the model available on torchhub.
At test time, to extract the landmark locations from the output heatmaps, we first filtered the outputs using a threshold value of 0.3, as per \cite{Duffy2021HighThroughputPP}.
The landmark location was then computed using the centroid of the pixels whose value was greater than 0.3. 

\subsection{Automated EF estimation application}

\textbf{Dataset setup and spatial-temporal model workflow.} We used the same dataset EchoNet-Dynamic and CAMUS dataset as in the structure segmentation. The ejection fraction (EF) values in these datasets were estimated from manual tracings of the left ventricle in the identified end-diastole (ED) and end-systole (ES) frames. The EchoNet-Dynamic videos consist of multiple heartbeats, while the CAMUS videos contain only a single heartbeat. 

Our model is designed to estimate the EF from a sequence of echocardiogram images in two stages: the first stage identifies the ED and ES frames, and the second stage estimates the EF from the sequence of images between these frames. In the first stage, we trained an image-to-image structure segmentation model using the pretrained EchoApex encoder for left ventricle segmentation. Following the approach in \cite{Ouyang2020VideobasedAF}, we determined the left ventricle volume from the segmentations and identified the ED and ES frames via peak finding from the volume size trajectory. In the second stage, we sampled frames between the ED and ES frames and directly regressed the EF value. Specifically, 8 frames were sampled between the ED and ES frames and independently processed by the EchoApex encoder to obtain feature embeddings. These embeddings were then combined with trainable temporal tokens, which encoded the temporal position of each frame and added temporal context to the spatial embeddings. We followed the classification task setup analogy by combining an EF prediction token with the embedded features and temporal tokens and forwarding them into a sequence decoder consisting of two transformer blocks. The decoder processed these integrated features to predict a single continuous value representing the EF. A Sigmoid final layer mapped the decoder's prediction into the (0,1) range, and a multiplication factor of 100 scaled it to (0,100). During the training of the second stage, we randomly sampled cardiac cycles from the training split of the EchoNet-Dynamic dataset and used a mean square error loss between the predicted EF value and the ground truth. The model was trained using the AdamW optimizer with a learning rate of 5e-4 and a weight decay of 0.005. 

\textbf{Evaluation of EchoApex on EF prediction.} We evaluated the performance of our model on both the EchoNet-Dynamic and CAMUS datasets. While sophisticated spatial-temporal models and learning strategies have been proposed for EF estimation in existing works \cite{muhtaseb2022echocotr, dai2022cyclical,alven2024deep}, we focused on the evaluation of the feature embeddings from the pretrained EchoApex encoder. Incorporating EchoApex into the advanced model frameworks is feasible but is beyond the scope of this work. We therefore compared the performance the EchoApex based model with the ImageNet pretrained counterpart. Throughout the training and evaluation, we frozen the image encoding part of the model and only trained the decoder part.
\subsection{Statistical analysis}
To compute the standard deviation in Fig. \ref{fig:view_cls_bar_and_table} and Fig. \ref{fig:view_cls_gridplot} of the view classification experiment, we perform nonparametric bootstrapping using 1000 samples and compute the balanced accuracy using the bootstrapped data. To compare the models' performance in the classification task, we use a nonparametric one-sided permutation test on the bootstrapped data, where the observations are randomly paired between the two samples.  The standard deviation in Fig. \ref{fig:dinosam_unet_medsam} was directly estimated from the model predictions, as Dice scores are continuous. For hypothesis testing of superior performance between model comparisons, including comparisons between different experiments in structure segmentation, left ventricle measurement, and automated ejection fraction estimation, we used a one-sided paired t-test. The p-values were computed using the t-distribution, with degrees of freedom equal to the number of samples minus one. The default significance level was set to 0.01.

\subsection{Software, hardware, and code}
We used PyDICOM and internally developed C\texttt{++} based tool for 2D and 3D data proeprocessing. We used Python 3.10 with Pytorch 2.1.0 for all the remaining pieces of model development and results analysis in the study. Model pretraining and finetuning for all tasks are performed on nodes with 8 80GB  NVIDIA H100 or A100 GPUs. Pretraining of EchoApex vitB model takes approximately 2100 H100 GPU hours. For pretraining, we referred the implementation of officially released DINOv2 repository (\url{https://github.com/facebookresearch/dinov2}) and re-implemented it into a framework with pytorch-lightning 2.1.2 (\url{https://github.com/Lightning-AI/pytorch-lightning/}) and lightly 1.4.23 (\url{https://github.com/lightly-ai/lightly}). The statistical tests were performed using the Python library scipy 1.7.3. The statistics calculation and summary are performed using Pandas 2.0.3.  The plots are generated using the Python library Matplotlib 3.4.3 and Seaborn 0.13.2.

\clearpage
\section{Supplementary Materials: Tables and Figures}

\subsection{Supp. of Pretraining}
\begin{table}[H]
    \centering
    \begin{tabular}{|l|l|l|}
    \hline
    Module                         & Hyperparameters     & Value       \\ \hline
    \multirow{5}{*}{Network ViT-B} & blocks              & 12          \\
                                   & heads               & 12          \\
                                   & embedding dim       & 768         \\
                                   & no. tokens          & 64          \\
                                   & patch size          & 14          \\ \hline
    \multirow{5}{*}{Augmentation}  & global crop size    & 112         \\
                                   & global crop scale   & (0.25, 1.0) \\
                                   & local crop size     & 42          \\
                                   & local crop scale    & (0.05, 0.25) \\
                                   & random mask ratio   & (0.1, 0.5)  \\ \hline
    \multirow{8}{*}{DINOv1}        & ts hidden dim       & 2048        \\
                                   & ts bottleneck dim   & 64          \\
                                   & ts output dim       & 65536       \\
                                   & warmup teacher temp & 0.04        \\
                                   & teacher temp        & 0.07        \\
                                   & optimizer           & AdamW       \\
                                   & learning rate       & 0.0005      \\
                                   & weight decay        & 0.005       \\ \hline
    \multirow{4}{*}{DINOv2}        & ibot hidden dim     & 2048        \\
                                   & ibot output dim     & 65536       \\
                                   & ibot seperate head  & true        \\
                                   & ibot loss weight    & 1.0         \\ \hline
    \end{tabular}
    \caption{\textbf{Hyperparameters of EchoApex with ViT-B encoder.} For the training efficiency, we leverage a two-stage approach: initialized from ImageNet pretrained model, we first pretrain the network with DINOv1 to convergence, and then continue training with DINOv2. A single node of $8 \times 80$GB Nvidia GPUs is used to train EchoApex-B, taking a total number of 2100 GPU hours for the pretraining on Echo20M dataset. }\label{tab:pretraining_hyperparameters}
    \end{table}
\subsection{Supp. of View Classification}
All the images were annotated by certified echocardiographers. Below, we detail the classes' abbreviations used in the figures and their meaning.

\begin{itemize}
    \item \textbf{A2C}: Apical two-chamber view.
    \item \textbf{A3C}: Apical three-chamber view.
    \item \textbf{A4C}: Apical four-chamber view.
    \item \textbf{A5C}: Apical five-chamber view.
    \item \textbf{Cont:A2C}: Apical two-chamber view with contrast.
    \item \textbf{Cont:A3C}: Apical three-chamber view with contrast.
    \item \textbf{Cont:A4C}: Apical two-chamber view with contrast.
    \item \textbf{Cont:SAX}: Parasternal short-axis view with contrast.
    \item \textbf{PLAX:ID}: Parasternal long-axis view with increased depth.
    \item \textbf{PLAX:LV}: Parasternal long-axis left ventricle .
    \item \textbf{PLAX:RVN}: Parasternal long-axis right ventricle inflow.
    \item \textbf{PLAX:RVT}: Parasternal long-axis right ventricle outflow.
    \item \textbf{PLAX:VAL}: Parasternal long-axis zoomed on the mitral and/or aortic valve.
    \item \textbf{PSAX:AV}: Parasternal short-axis with a focus on the aortic valve.
    \item \textbf{PSAX:MV}: Parasternal short-axis with a focus on the mitral valve.
    \item \textbf{PSAX:PAP}: Parasternal short-axis at the papillary muscles level.
    \item \textbf{SC:4C}: Subcostal four-chamber view.
    \item \textbf{SC:IVC}: Subcostal long axis inferior vena cava view.
\end{itemize}

\begin{figure}[h]
     \centering
     \includegraphics[width=0.8\textwidth]{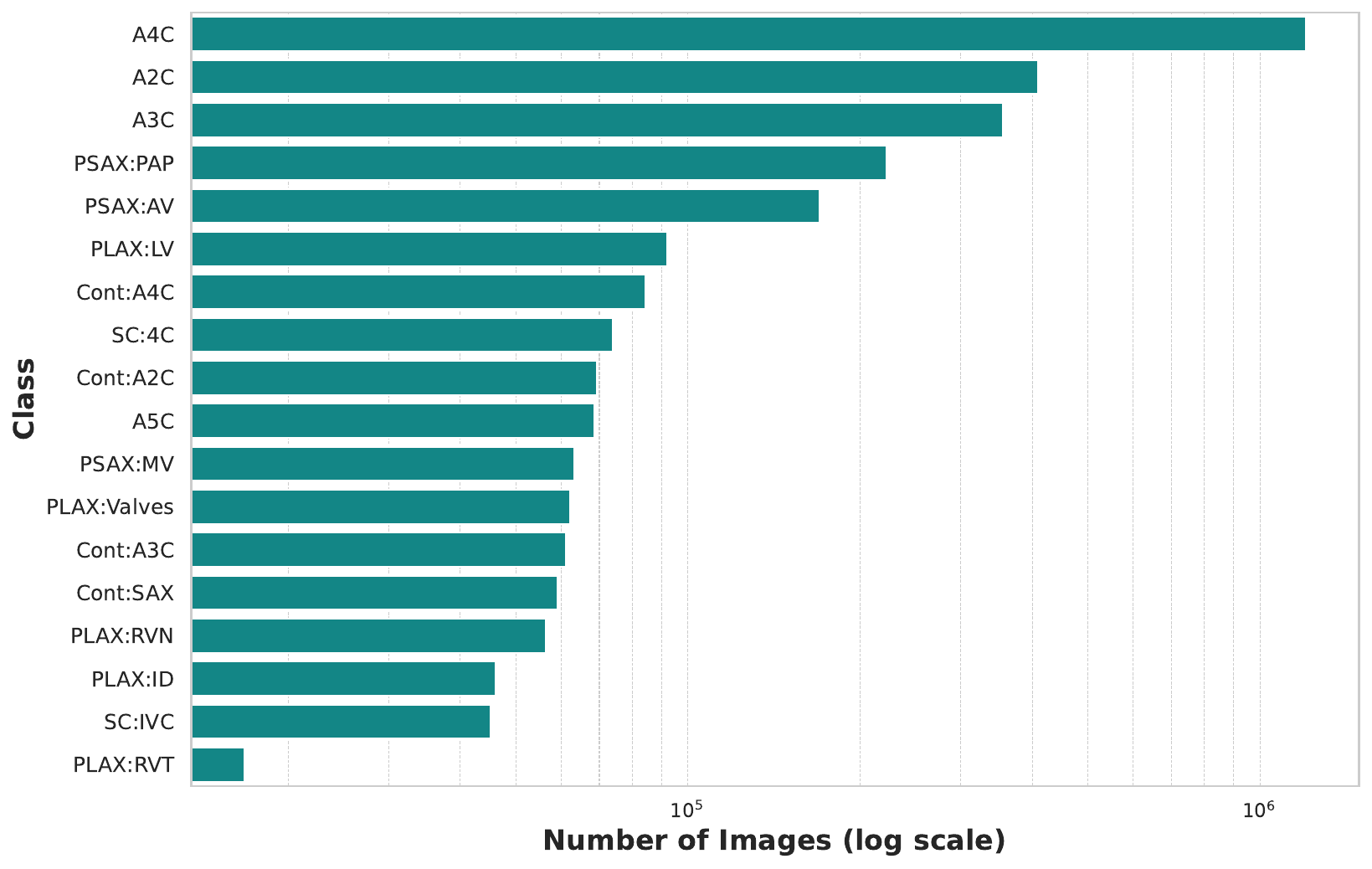}
     \caption{\textbf{Number of images per TTE class for the view classification task.}}
     \label{fig:supp_view_cls_distribution}
\end{figure}

In Fig. \ref{fig:supp_view_cls_distribution}, we show the distribution of the TTE classes for the whole dataset.

\begin{table}[h!]
    \centering
    \begin{tabular}{|c|c|c|c|c|c|}
    
    \hline
     Backbone & Model & Parameter & Value  \\
     \hline
     \multirow{2}{*}{ResNet50-IN} & \multirow{2}{*}{ResNet50-IN} & Optimizer & AdamW \\
        &  & Learning rate & 1.0e-4 \\
        &  & Batch size & 256 \\
      \hline
     \multirow{3}{*}{ViT-S} & \multirow{3}{*}{EchoApex} & Optimizer & AdamW \\
                            &  & Learning rate & 1.0e-4 \\
                            &  & Batch size & 256 \\
                            &  & Adapter hidden size & 192 \\
      \hline
     \multirow{3}{*}{ViT-B} & \multirow{3}{*}{EchoApex} & Optimizer & AdamW \\
                            &  & Learning rate & 1.0e-5 \\
                            &  & Batch size & 32 \\
                            &  & Adapter hidden size & 384 \\
      \hline
     \end{tabular}
     \caption{\textbf{Hyperparameters used for the view classification experiment}. Batch size refers to the batch size per GPU.}
      \label{table:supp_view_cls_hyperparams}
    \end{table}
    
    \begin{table}[h!]
    \centering
    \begin{tabular}{|c|c|c|c|c|c|}
    
    \hline
     Backbone & Model & Adapters & Accuracy & F1-score & BACC \\
     \hline
     Resnet50 & ResNet50-IN & \XSolidBrush & 0.923 & 0.917 & 0.959 \\
      \hline
     \multirow{2}{*}{ViT-S} &  \multirow{2}{*}{EchoApex} & \Checkmark & 0.940 & 0.942 & 0.968 \\
      &  & \XSolidBrush & 0.948 & 0.948 & 0.972  \\
      \hline
     \multirow{2}{*}{ViT-B} & \multirow{2}{*}{EchoApex} & \Checkmark & 0.954 & 0.955 & 0.975 \\
    &  & \XSolidBrush & 0.956 & 0.954 & 0.976 \\
      \hline
     \end{tabular}
     \caption{\textbf{Model performance for TTE view classification task, averaged over the 18 classes.} BACC: Balanced Accuracy}
      \label{table:supp_view_cls_results}
    \end{table}
    
    
    \begin{table}[h!]
    \centering
    \begin{tabular}{|c|c|c|p{1.5cm}|p{1.5cm}|p{1.5cm}|p{1.5cm}|}
    
    \hline
     Backbone & Model & Adapters & Accuracy [EchoNet] & Accuracy [CAMUS] & F1-score [CAMUS] & BACC [CAMUS] \\
     \hline
        Resnet50 & ResNet50-IN & \XSolidBrush & 0.764 & 0.902 & 0.931 & 0.935 \\
      \hline
      \multirow{2}{*}{ViT-S} &  \multirow{2}{*}{EchoApex} & \Checkmark & 0.922 & 0.942 & 0.961 & 0.962 \\
        &  & \XSolidBrush & 0.853  & 0.947 & 0.962 & 0.963  \\
      \hline
      \multirow{2}{*}{ViT-B} & \multirow{2}{*}{EchoApex} & \Checkmark & 0.832 & 0.926 & 0.955 & 0.957 \\
        &  & \XSolidBrush & 0.874 & 0.939 & 0.962 & 0.963 \\
      \hline
     \end{tabular}
     \caption{\textbf{Model performance on the TTE view classification task on two unseen datasets, CAMUS and EchoNet-Dynamic}. A2C and A4C images are present in the CAMUS dataset and only A4C images are available in the EchoNet-Dynamic dataset. We only report accuracy for the EchoNet dataset.}
      \label{table:zero_shot_view_cls_results}
    \end{table}
    
    
    \begin{table}[t!]
    \centering
    \begin{tabular}{|c|c|c|c|c|c|}
    \hline
         Backbone & Model & Class &  Accuracy &  F1-Score &  BACC \\
    \hline
    
    \multirow{18}{*}{ResNet50} & \multirow{18}{*}{ResNet50} & A2C & 0.982 & 0.985 & 0.991 \\
     & & A3C & 1.000 & 0.889 & 0.99 \\
     & & A4C & 0.992 & 0.99 & 0.995 \\
     & & A5C & 0.952 & 0.968 & 0.976 \\
     & & Cont:A2C & 0.923 & 0.835 & 0.958 \\
     & & Cont:A3C & 0.889 & 0.923 & 0.944 \\
     & & Cont:A4C & 0.892 & 0.938 & 0.946 \\
     & & Cont:SAX & 1.000 & 0.987 & 1.000 \\
     & & PLAX:LV & 0.802 & 0.838 & 0.894 \\
     & & PLAX:ID & 0.690 & 0.813 & 0.845 \\
     & & PLAX:RVN & 0.936 & 0.946 & 0.967 \\
     & & PLAX:RVT & 0.824 & 0.848 & 0.911 \\
     & & PLAX:Valves & 0.909 & 0.866 & 0.949 \\
     & & PSAX:AV & 1.000 & 0.884 & 0.996 \\
     & & PSAX:PAP & 0.946 & 0.96 & 0.971 \\
     & & PSAX:MV & 0.883 & 0.866 & 0.939 \\
     & & SC:4C & 1.000 & 0.99 & 1.000 \\
     & & SC:IVC & 1.000 & 0.977 & 1.000 \\

    \hline
    \end{tabular}
    \caption{\textbf{Detailed performance metrics for the ResNet50 model for the view classfication task.}}
    \label{tab:supp_view_cls_resnet_detailed_results}
    \end{table}
    

    \clearpage
    \begin{table}[t!]
    \centering
    \begin{tabular}{|c|c|c|c|c|c|}
    \hline
         Backbone & Model & Class &  Accuracy &  F1-Score &  BACC \\
    \hline
    
    \multirow{18}{*}{ViT-S} & \multirow{18}{*}{EchoApex (Adapters)} & A2C & 0.982 & 0.985 & 0.991 \\
     & & A3C & 1.000 & 0.948 & 0.996 \\
     & & A4C & 1.000 & 0.995 & 0.999 \\
     & & A5C & 0.952 & 0.968 & 0.976 \\
     & & Cont:A2C & 0.942 & 0.845 & 0.968 \\
     & & Cont:A3C & 0.907 & 0.933 & 0.953 \\
     & & Cont:A4C & 0.860 & 0.914 & 0.930 \\
     & & Cont:SAX & 1.000 & 0.987 & 1.000 \\
     & & PLAX:LV & 0.942 & 0.892 & 0.961 \\
     & & PLAX:ID & 0.721 & 0.830 & 0.860 \\
     & & PLAX:RVN & 0.936 & 0.967 & 0.968 \\
     & & PLAX:RVT & 1.000 & 0.971 & 1.000 \\
     & & PLAX:Valves & 0.868 & 0.890 & 0.931 \\
     & & PSAX:AV & 1.000 & 1.000 & 1.000 \\
     & & PSAX:PAP & 0.984 & 0.972 & 0.989 \\
     & & PSAX:MV & 0.831 & 0.883 & 0.915 \\
     & & SC:4C & 1.000 & 0.990 & 1.000 \\
     & & SC:IVC & 1.000 & 1.000 & 1.000 \\
    
    \hline
    
    \multirow{18}{*}{ViT-S} &  \multirow{18}{*}{EchoApex} & A2C & 0.988 & 0.988 & 0.994 \\
     & & A3C & 1.000 & 0.943 & 0.995 \\
     & & A4C & 0.997 & 0.995 & 0.998 \\
     & & A5C & 0.968 & 0.984 & 0.984 \\
     & & Cont:A2C & 0.962 & 0.877 & 0.978 \\
     & & Cont:A3C & 0.889 & 0.941 & 0.944 \\
     & & Cont:A4C & 0.925 & 0.950 & 0.962 \\
     & & Cont:SAX & 0.987 & 0.981 & 0.993 \\
     & & PLAX:LV & 0.888 & 0.881 & 0.936 \\
     & & PLAX:ID & 0.729 & 0.828 & 0.863 \\
     & & PLAX:RVN & 0.947 & 0.973 & 0.973 \\
     & & PLAX:RVT & 1.000 & 0.971 & 1.000 \\
     & & PLAX:Valves & 0.926 & 0.885 & 0.958 \\
     & & PSAX:AV & 1.000 & 1.000 & 1.000 \\
     & & PSAX:PAP & 0.984 & 0.978 & 0.990 \\
     & & PSAX:MV & 0.883 & 0.907 & 0.940 \\
     & & SC4C & 1.000 & 0.990 & 1.000 \\
     & & SC:IVC & 1.000 & 1.000 & 1.000 \\
    
    \hline
    \end{tabular}
    \caption{\textbf{Detailed performance metrics of the EchoApex model with ViT-S backbone on the view classification task.}}
    \label{tab:supp_view_cls_vitS_detailed_results}
    \end{table}
    
    
    \begin{table}[H]
    \centering
    \begin{tabular}{|c|c|c|c|c|c|}
    \hline
         Backbone & Model & Class &  Accuracy &  F1-Score &  BACC \\
    \hline
    
    \multirow{18}{*}{ViT-B} & \multirow{18}{*}{EchoApex (Adapters)} & A2C & 0.988 & 0.988 & 0.994 \\
     & & A3C & 1.000 & 0.96 & 0.997 \\
     & & A4C & 1.000 & 0.997 & 0.999 \\
     & & A5C & 1.000 & 1.000 & 1.000 \\
     & & Cont:A2C & 0.923 & 0.881 & 0.959 \\
     & & Cont:A3C & 0.944 & 0.971 & 0.972 \\
     & & Cont:A4C & 0.925 & 0.945 & 0.962 \\
     & & Cont:SAX & 1.000 & 0.987 & 1.000 \\
     & & PLAX:LV & 0.909 & 0.896 & 0.947 \\
     & & PLAX:ID & 0.806 & 0.86 & 0.901 \\
     & & PLAX:RVN & 0.947 & 0.973 & 0.973 \\
     & & PLAX:RVT & 1.000 & 0.971 & 1.000 \\
     & & PLAX:Valves & 0.893 & 0.900 & 0.944 \\
     & & PSAX:AV & 1.000 & 1.000 & 1.000 \\
     & & PSAX:PAP & 0.975 & 0.972 & 0.985 \\
     & & PSAX:MV & 0.870 & 0.882 & 0.933 \\
     & & SC:4C & 1.000 & 1.000 & 1.000 \\
     & & SC:IVC & 1.000 & 1.000 & 1.000 \\
    
    \hline   
    \multirow{18}{*}{ViT-B} & \multirow{18}{*}{EchoApex} & A2C & 0.982 & 0.988 & 0.991 \\
     & & A3C & 1.000 & 0.926 & 0.994 \\
     & & A4C & 0.997 & 0.994 & 0.998 \\
     & & A5C & 0.984 & 0.984 & 0.992 \\
     & & Cont:A2C & 0.942 & 0.916 & 0.970 \\
     & & Cont:A3C & 0.963 & 0.963 & 0.981 \\
     & & Cont:A4C & 0.946 & 0.967 & 0.973 \\
     & & Cont:SAX & 1.000 & 0.994 & 1.000 \\
     & & PLAX:LV & 0.897 & 0.889 & 0.941 \\
     & & PLAX:ID & 0.791 & 0.864 & 0.894 \\
     & & PLAX:RVN & 0.947 & 0.973 & 0.973 \\
     & & PLAX:RVT & 1.000 & 0.971 & 1.000 \\
     & & PLAX:Valves & 0.884 & 0.899 & 0.94 \\
     & & PSAX:AV & 1.000 & 1.000 & 1.000 \\
     & & PSAX:PAP & 0.959 & 0.97 & 0.978 \\
     & & PSAX:MV & 0.922 & 0.882 & 0.958 \\
     & & SC:4C & 1.000 & 1.000 & 1.000 \\
     & & SC:IVC & 1.000 & 1.000 & 1.000 \\
     
    \hline
    \end{tabular}
    \caption{\textbf{Detailed performance metrics of the EchoApex model with ViT-B backbone on the view classification task.}}
    \label{tab:supp_view_cls_vitB_detailed_results}
    \end{table}
    
    
    
    \begin{figure}[h!]
         \centering
         \begin{subfigure}{0.19\textwidth}
             \centering
             \includegraphics[width=\textwidth]{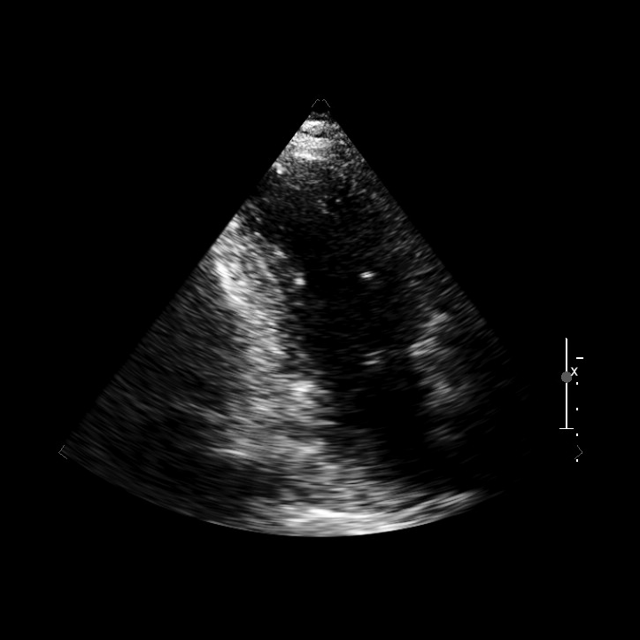}
             \caption{A2C}
         \end{subfigure}
         \begin{subfigure}{0.19\textwidth}
             \centering
             \includegraphics[width=\textwidth]{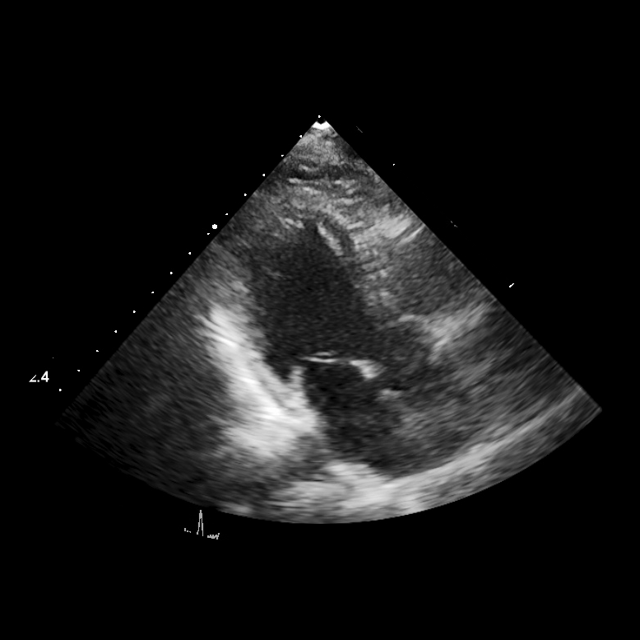}
             \caption{A3C}
         \end{subfigure}
         \begin{subfigure}{0.19\textwidth}
             \centering
             \includegraphics[width=\textwidth]{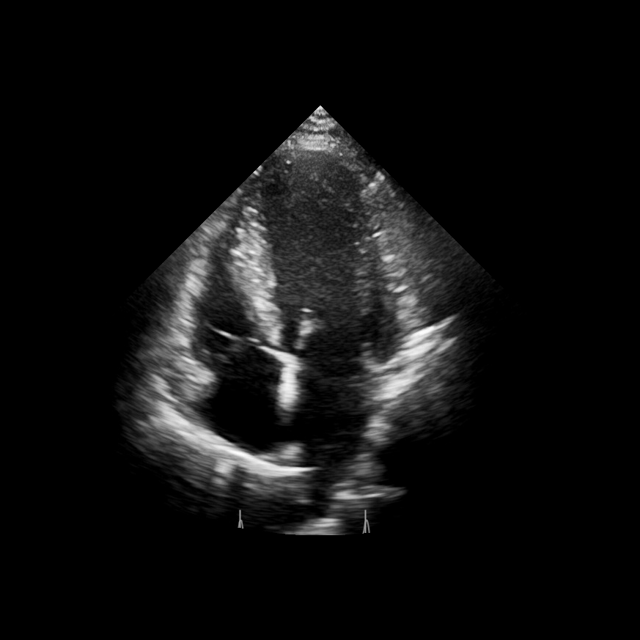}
             \caption{A4C}
         \end{subfigure}
         \begin{subfigure}{0.19\textwidth}
             \centering
             \includegraphics[width=\textwidth]{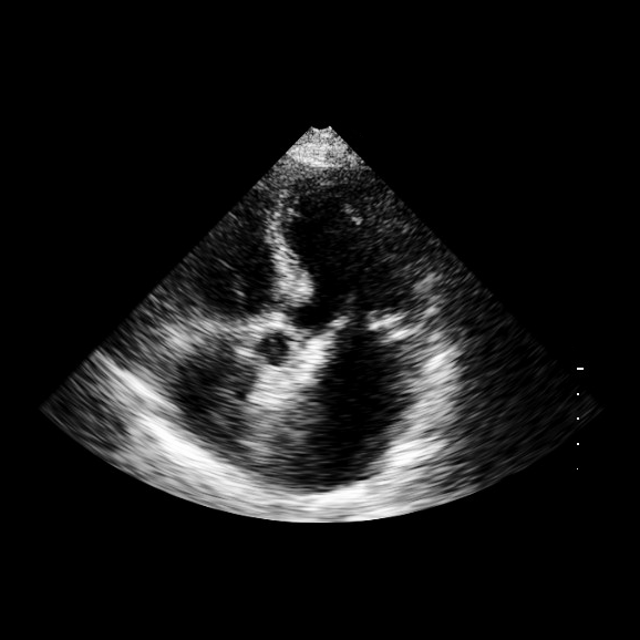}
             \caption{A5C}
         \end{subfigure}
         \begin{subfigure}{0.19\textwidth}
             \centering
             \includegraphics[width=\textwidth]{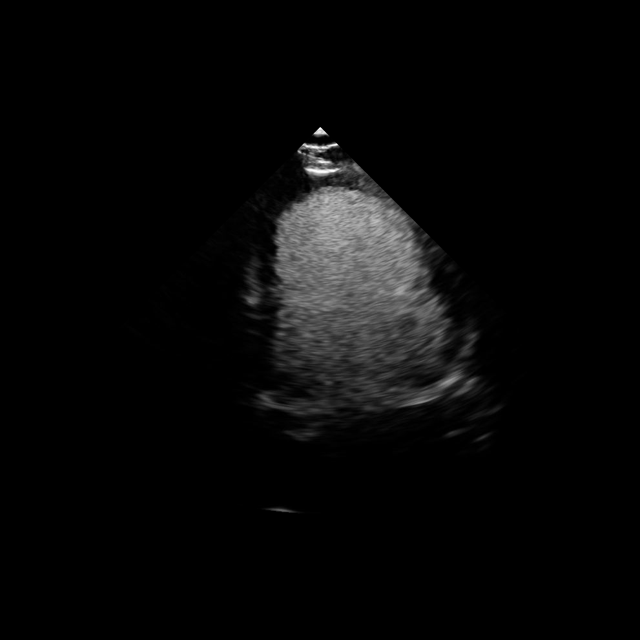}
             \caption{Cont:A2C}
         \end{subfigure}
    
         \begin{subfigure}{0.19\textwidth}
             \centering
             \includegraphics[width=\textwidth]{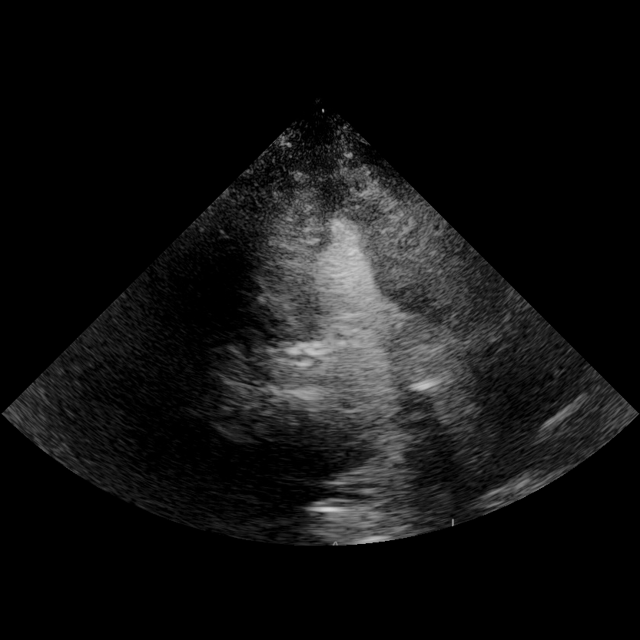}
             \caption{Cont:A3C}
         \end{subfigure}
         \begin{subfigure}{0.19\textwidth}
             \centering
             \includegraphics[width=\textwidth]{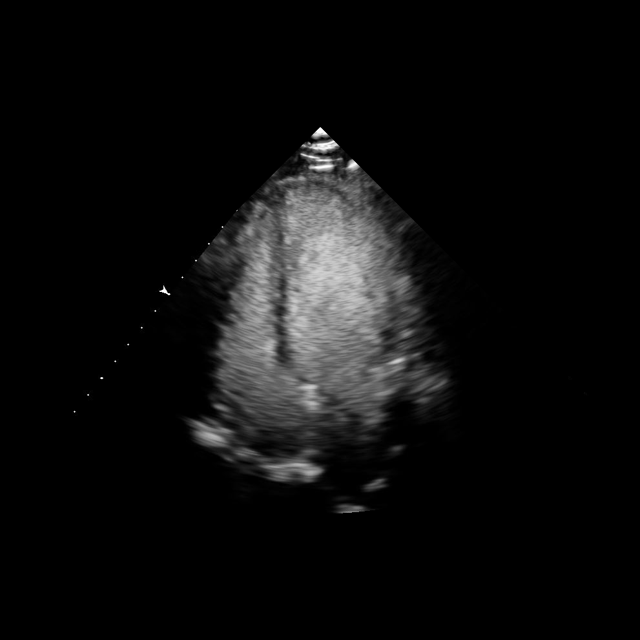}
             \caption{Cont:A4C}
         \end{subfigure}
         \begin{subfigure}{0.19\textwidth}
             \centering
             \includegraphics[width=\textwidth]{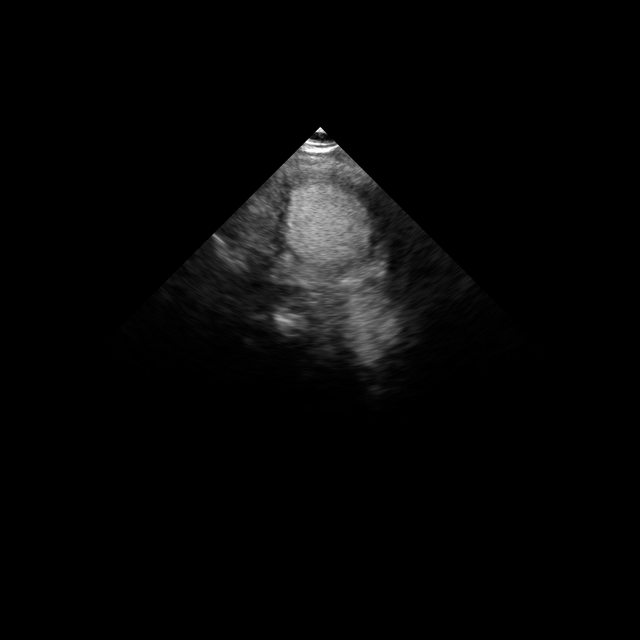}
             \caption{Cont:SAX}
         \end{subfigure}
          \begin{subfigure}{0.19\textwidth}         \centering
             \includegraphics[width=\textwidth]{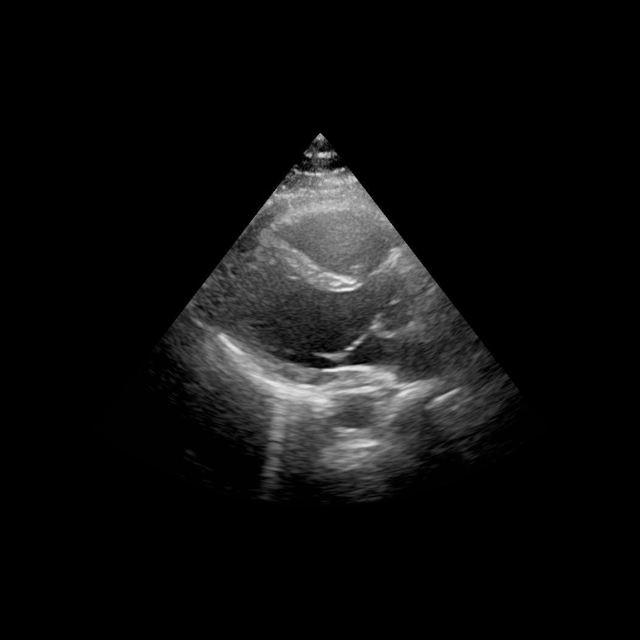}
             \caption{PLAX:ID}
         \end{subfigure}
         \begin{subfigure}{0.19\textwidth}
             \centering
             \includegraphics[width=\textwidth]{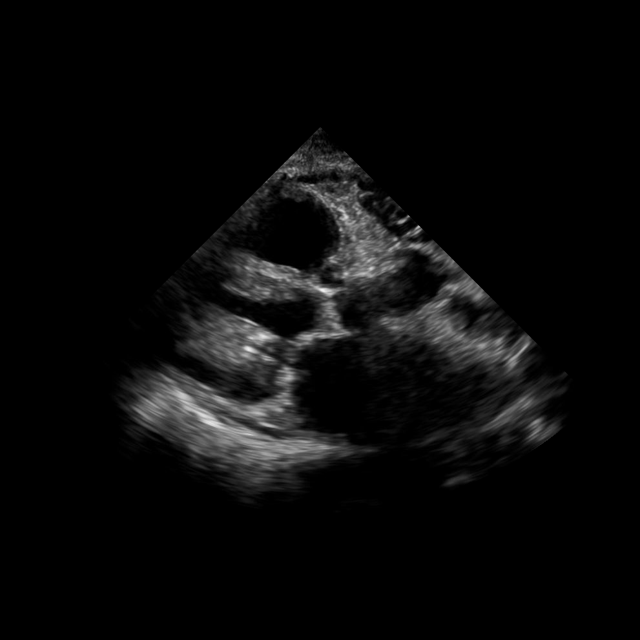}
             \caption{PLAX:LV}
         \end{subfigure}
    
         \begin{subfigure}{0.19\textwidth}
             \centering
             \includegraphics[width=\textwidth]{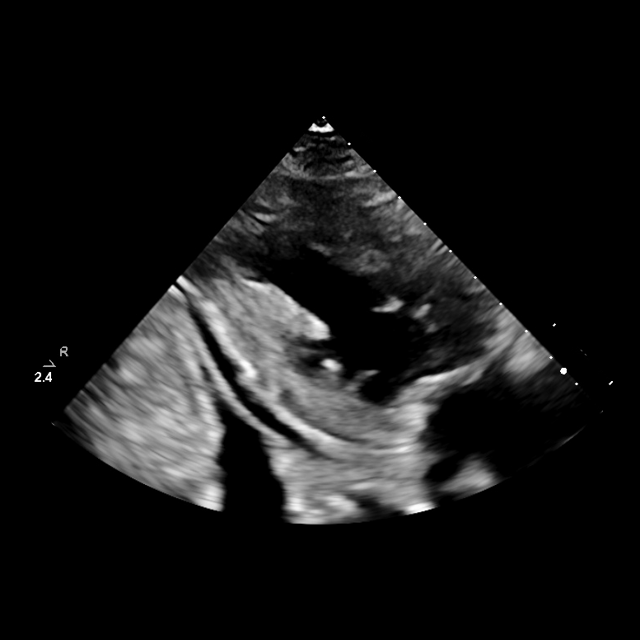}
             \caption{PLAX:RVN}
         \end{subfigure}
         \begin{subfigure}{0.19\textwidth}
             \centering
             \includegraphics[width=\textwidth]{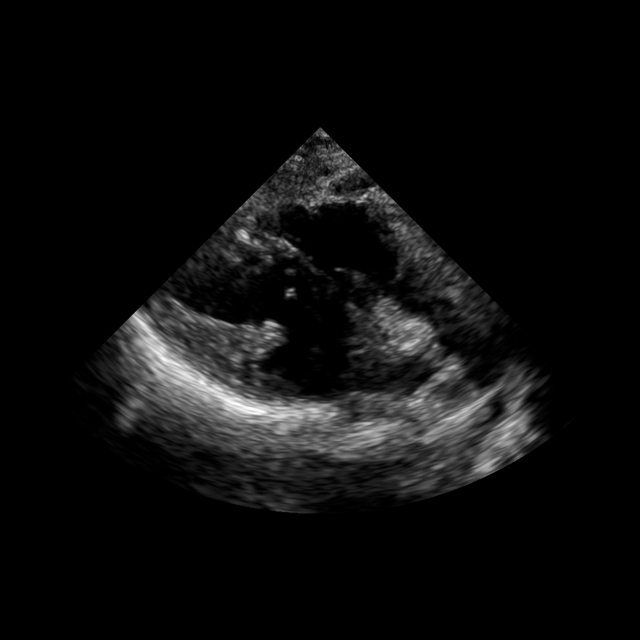}
             \caption{PLAX:RVT}
         \end{subfigure}
         \begin{subfigure}{0.19\textwidth}
             \centering
             \includegraphics[width=\textwidth]{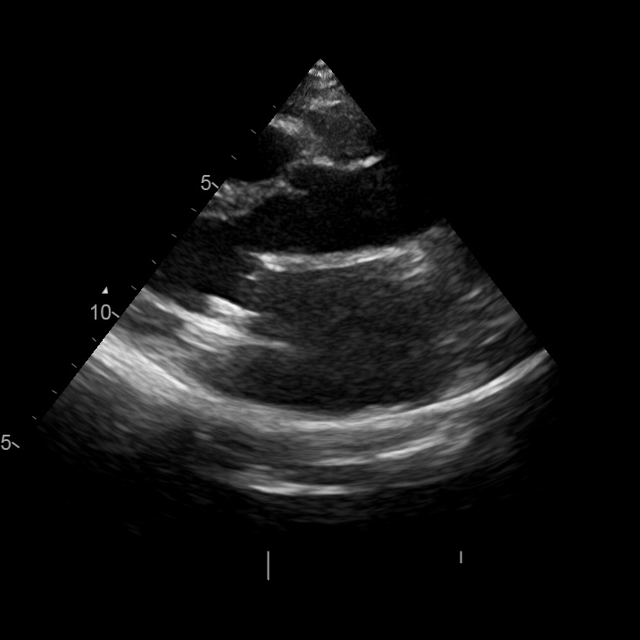}
             \caption{PLAX:VAL}
         \end{subfigure}
         \begin{subfigure}{0.19\textwidth}
             \centering
             \includegraphics[width=\textwidth]{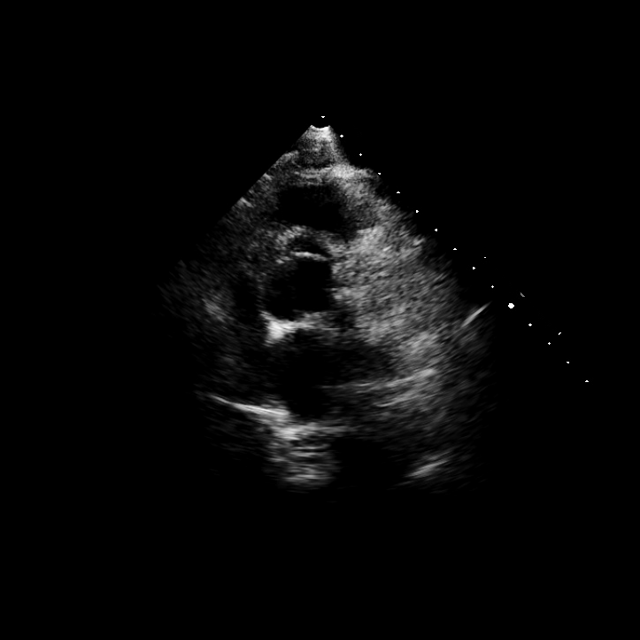}
             \caption{PSAX:AV}
         \end{subfigure}
         \begin{subfigure}{0.19\textwidth}
             \centering
             \includegraphics[width=\textwidth]{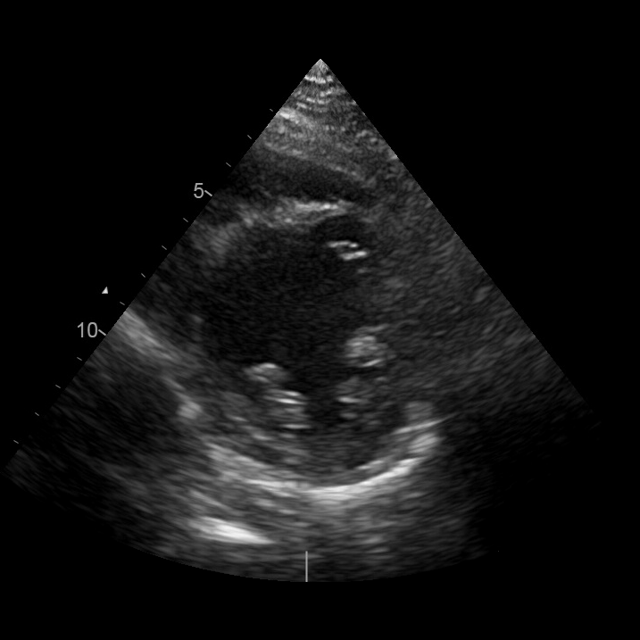}
             \caption{PSAX:PAP}
         \end{subfigure}
    
         \begin{subfigure}{0.19\textwidth}
             \centering
             \includegraphics[width=\textwidth]{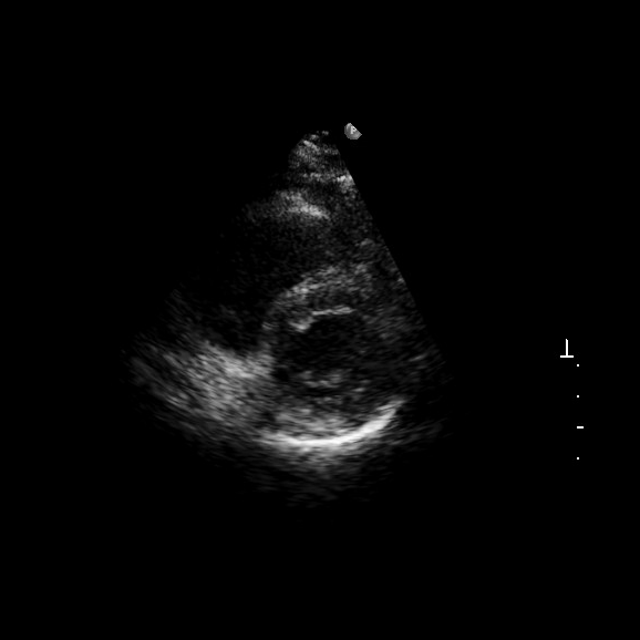}
             \caption{PSAX:MV}
         \end{subfigure}
         \begin{subfigure}{0.19\textwidth}
             \centering
             \includegraphics[width=\textwidth]{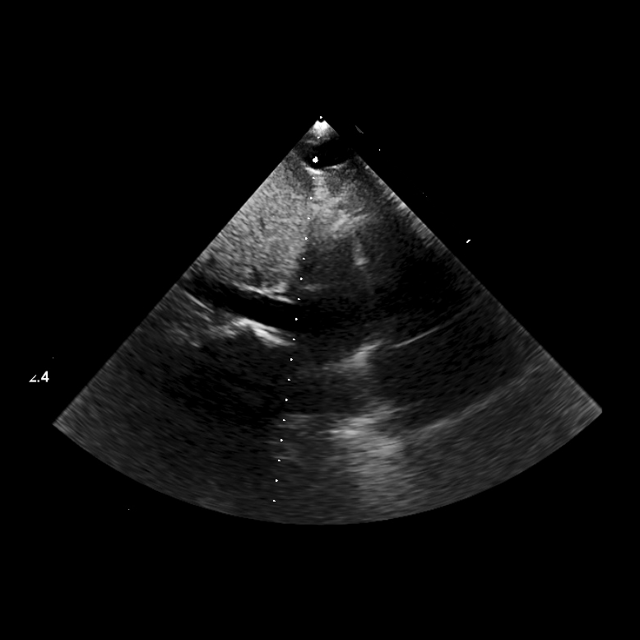}
             \caption{SC:IVC}
         \end{subfigure}
         \begin{subfigure}{0.19\textwidth}
             \centering
             \includegraphics[width=\textwidth]{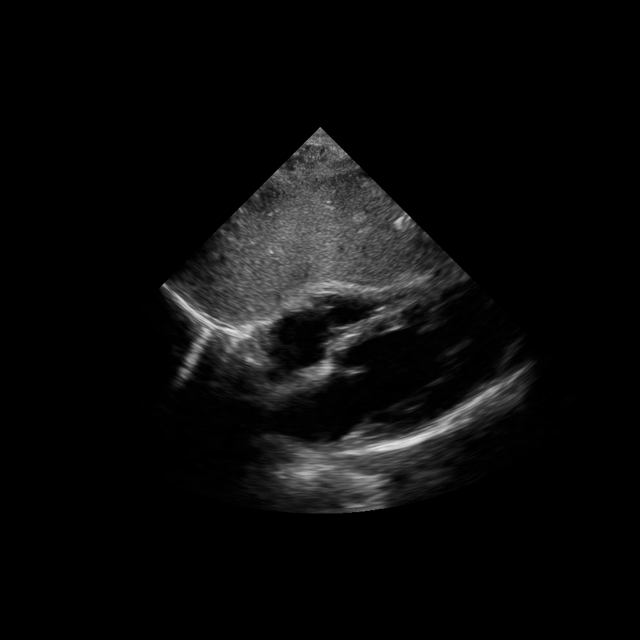}
             \caption{SC:4C}
         \end{subfigure}
    
        \caption{\textbf{Example of images used in the TTE view classification task.}}
        \label{fig:supp_view_cls_gallery}
    \end{figure}
    
    
\subsection{Supp. of Structure Segmentation}

\begin{table}[H]
\begin{tabular}{|c|c|c|c|c|c|c|}
\hline
Dataset                            & Structure           & Views & Patients              & Videos & Annotation & \begin{tabular}[c]{@{}c@{}}Evaluation \\ (ED, ES)\end{tabular} \\ \hline
EchoNet-Dynamic                    & LV                  & A4C   & 10030                 & 10030  & 20048      & 2552                                                           \\ \hline
\multirow{4}{*}{CAMUS}             & \multirow{2}{*}{LV} & A4C   & \multirow{4}{*}{500}  & 500    & 9964       & 80                                                             \\ \cline{3-3} \cline{5-7} 
                                   &                     & A2C   &                       & 500    & 9268       & 80                                                             \\ \cline{2-3} \cline{5-7} 
                                   & \multirow{2}{*}{LA} & A4C   &                       & 500    & 9964       & 80                                                             \\ \cline{3-3} \cline{5-7} 
                                   &                     & A2C   &                       & 500    & 9264       & 80                                                             \\ \hline
Vol-biplane                        & four chambers       & -     & 1231                  & 1231   & 9129       & 1827                                                           \\ \hline
\multirow{2}{*}{EchoNet-Pediatric} & LV                  & A4C   & \multirow{2}{*}{1958} & 3176   & 6449       & 1386                                                           \\ \cline{2-3} \cline{5-7} 
                                   & LV                  & PSAX  &                       & 4424   & 9001       & 1928                                                           \\ \hline
\end{tabular} \caption{\textbf{Details of dataset for structure segmentation.} We used three publicly available dataset and their official training/validation/testing split for the model assessment. We also added an internal dataset consists of biplane annotations in volumetric TEE and TTE echo. } \label{table:interactivesegmentation_datadetails}
\end{table}

\begin{table}[H]
	\begin{tabular}{|c|c|c|c|cc|}
	\hline
	 &  &  &  & \multicolumn{2}{c|}{EchoApex-B} \\ \cline{5-6} 
	\multirow{-2}{*}{Dataset} & \multirow{-2}{*}{Target} & \multirow{-2}{*}{Specialist} & \multirow{-2}{*}{MedSAM} & \multicolumn{1}{c|}{EchoApex-SAM-AdaB} & EchoApex-SAM-B \\ \hline
	CAMUS & LV-ES & 0.916 & 0.856 & \multicolumn{1}{c|}{0.920} & 0.930 \\ \hline
	 & LV-ED & 0.939 & 0.887 & \multicolumn{1}{c|}{0.937} & 0.945 \\ \hline
	\rowcolor[HTML]{EFEFEF} 
	CAMUS LV mean &  & 0.928 & 0.872 & \multicolumn{1}{c|}{\cellcolor[HTML]{EFEFEF}0.929} & 0.938 \\ \hline
	 & LA-ES & 0.918 & 0.825 & \multicolumn{1}{c|}{0.901} & 0.920 \\ \hline
	 & LA-ED & 0.889 & 0.780 & \multicolumn{1}{c|}{0.874} & 0.896 \\ \hline
	\rowcolor[HTML]{EFEFEF} 
	CAMUS LA mean &  & 0.904 & 0.803 & \multicolumn{1}{c|}{\cellcolor[HTML]{EFEFEF}0.888} & 0.908 \\ \hline
	ENDym & LV-ES & 0.903 & 0.845 & \multicolumn{1}{c|}{0.910} & 0.917 \\ \hline
	 & LV-ED & 0.927 & 0.885 & \multicolumn{1}{c|}{0.930} & 0.939 \\ \hline
	\rowcolor[HTML]{EFEFEF} 
	ENDym LV mean &  & 0.915 & 0.865 & \multicolumn{1}{c|}{\cellcolor[HTML]{EFEFEF}0.920} & 0.928 \\ \hline
	ENPed-A4C & LV & 0.891 & 0.872 & \multicolumn{1}{c|}{0.913} & 0.921 \\ \hline
	ENPed-PSAX & LV & 0.896 & 0.880 & \multicolumn{1}{c|}{0.925} & 0.930 \\ \hline
	\end{tabular}
\caption{\textbf{Benchmark of segmentation performance with state-of-the-art models.} Comparison of task specialist model, medical generalist model MedSAM and EchoApex models on three large public echocardiogram datasets CAMUS, EchoNet-Dynamic (ENDym) and  EchoNet-Pediatric (ENPed) datasets. Metric represents average Dice Similarity Coefficient. Proposed EchoApex model shows improvement over task-specialist and medical generalist models, indicating the importance of model pretraining with in-domain data. } \label{table:interactivesegmentation_specialistmedsam}
\end{table}
\clearpage
\subsection{Supp. of Left Ventricle Measurements}
\begin{table}[h!]
    \centering
    \begin{tabular}{|c|c|c|c|c|c|}
    
    \hline
     Backbone & Model & Parameter & Value  \\
     \hline
     \multirow{2}{*}{DeepLabV3} & \multirow{2}{*}{DeepLabV3} & Optimizer & AdamW \\
      &  & Learning rate & 1.0e-3 \\
      &  & Gaussian ball std $\sigma$ & 1.0 \\
      & & Weighted MSE loss alpha & 1.0e-3 \\
      &  & Batch size & 32 \\
      \hline
     \multirow{4}{*}{ViT-S} & \multirow{4}{*}{EchoApex} & Optimizer & AdamW \\
                            &  & Learning rate & 1.0e-4 \\
                            &  & Focal loss $\gamma$ & 2.0 \\
                            &  & Gaussian ball std $\sigma$ & 5.0 \\
                            &  & Decoder num parameters & 3 M \\
                            &  & Decoder upsampling & ConvTranspose \\
                            &  & Decoder num upsampling blocks & 3 \\
                            &  & Adapter hidden size & 192 \\
                            & & Batch size & 16 \\
      \hline
     \multirow{4}{*}{ViT-B} & \multirow{4}{*}{EchoApex} & Optimizer & AdamW \\
                            &  & Learning rate & 1.0e-4 \\
                            &  & Focal loss $\gamma$ & 2.0 \\
                            &  & Gaussian ball std $\sigma$ & 5.0 \\
                            &  & Decoder num parameters & 14 M \\
                            &  & Decoder upsampling & ConvTranspose \\
                            &  & Decoder num upsampling blocks & 4 \\
                            &  & Adapter hidden size & 384 \\
                            & & Batch size & 8 \\
      \hline
     \end{tabular}
     \caption{\textbf{Hyperparameters used for the left ventricle measurements experiment.} Batch size refers to the batch size per GPU.}
      \label{table:supp_lvh_hyperparams}
    \end{table}
    
    \begin{table}[h!]
    \centering
    \begin{tabular}{|c|c|c|c|c|c|c|}
    
    \hline
     Backbone & Model & Adapters & \multicolumn{3}{|c|}{MAE} & $R^2$ \\
     \hline
      &  &  & IVS & LVID & LVPW &  \\
      &  &  & Mean $\pm$ std & Mean $\pm$ std & Mean $\pm$ std & \\
     \hline
     DeepLabV3 & DeepLabV3 & \XSolidBrush & 1.77 $\pm$ 2.04 & 3.23 $\pm$ 2.75  & 1.56 $\pm$ 1.30 & 0.95 \\
     \hline
     \multirow{2}{*}{ViT-S}& \multirow{2}{*}{EchoApex} & \Checkmark & 1.57 $\pm$ 1.64 & 2.57 $\pm$ 2.21 & 1.43 $\pm$ 1.15 & 0.96 \\
     & & \XSolidBrush & 1.81 $\pm$ 1.82 & 2.75 $\pm$ 2.37 & 1.61 $\pm$ 1.29 & 0.96  \\
     \hline
     \multirow{2}{*}{ViT-B}& \multirow{2}{*}{EchoApex} & \Checkmark & 1.90 $\pm$ 2.17 & 3.15 $\pm$ 3.10 & 1.92 $\pm$ 3.05 & 0.93 \\
     & & \XSolidBrush & 2.03 $\pm$ 2.70 & 3.06 $\pm$ 3.67 & 1.81 $\pm$ 2.24 & 0.94 \\
     \hline
     \end{tabular}
     \caption{\textbf{Model performance (MAE) on the internal EchoNet-LVH dataset.} MAE: Mean absolute error. }
     \label{table:supp_lvh_mae_results}
    \end{table}
    
    \begin{table}[h!]
    \centering
    \begin{tabular}{|c|c|c|c|c|c|}
    
    \hline
     Backbone & Model & Adapters & \multicolumn{3}{|c|}{Averaged landmark error} \\
     \hline
      &  &  & IVS & LVID & LVPW  \\
      &  &  & Mean $\pm$ std & Mean $\pm$ std & Mean $\pm$ std \\
     \hline
     DeepLabV3 & DeepLabV3 & \XSolidBrush & 4.68 $\pm$ 3.12 & 5.35 $\pm$ 3.09  & 5.94 $\pm$ 4.06 \\
     \hline
     \multirow{2}{*}{ViT-S}& \multirow{2}{*}{EchoApex} & \Checkmark & 4.10 $\pm$ 3.18 & 4.75 $\pm$ 3.30 & 5.63 $\pm$ 4.19 \\
     & & \XSolidBrush & 4.11 $\pm$ 3.10 & 4.89 $\pm$ 3.06 & 6.00 $\pm$ 4.89 \\
     \hline
     \multirow{2}{*}{ViT-B}& \multirow{2}{*}{EchoApex} & \Checkmark & 4.37 $\pm$ 3.43 & 5.19 $\pm$ 3.99 & 6.17 $\pm$ 4.55 \\
     & & \XSolidBrush & 4.73 $\pm$ 3.68 & 5.49 $\pm$ 3.67 & 6.99 $\pm$ 5.70 \\
     \hline
     \end{tabular}
     \caption{\textbf{Model performance (Averaged landmark error) on the internal EchoNet-LVH dataset.}}
     \label{table:supp_lvh_lmk_results}
    \end{table}
    
    \begin{table}[h!]
    \centering
    \begin{tabular}{|c|c|c|c|c|c|c|}
    
    \hline
     Backbone & Model & Adapters & \multicolumn{3}{|c|}{MAE} & $R^2$ \\
     \hline
      &  &  & IVS & LVID & LVPW &  \\
      &  &  & Mean $\pm$ std & Mean $\pm$ std & Mean $\pm$ std & \\
     \hline
     DeepLabV3 & DeepLabV3 & \XSolidBrush & 3.78 $\pm$ 3.90 & 7.54 $\pm$ 8.05  & 2.71 $\pm$ 3.18 & 0.72 \\
     \hline
     \multirow{2}{*}{ViT-S}& \multirow{2}{*}{EchoApex} & \Checkmark & 2.92 $\pm$ 4.44 & 5.60 $\pm$ 6.14 & 2.87 $\pm$ 3.63 & 0.81 \\
     & & \XSolidBrush & 2.68 $\pm$ 2.83 & 4.82 $\pm$ 5.50 & 2.50 $\pm$ 2.59 & 0.87  \\
     \hline
     \multirow{2}{*}{ViT-B}& \multirow{2}{*}{EchoApex} & \Checkmark & 3.24 $\pm$ 4.82 & 6.75 $\pm$ 8.76 & 3.17 $\pm$ 3.84 & 0.70 \\
     & & \XSolidBrush & 3.91 $\pm$ 5.15 & 6.09 $\pm$ 6.70 & 3.21 $\pm$ 4.90 & 0.74 \\
     \hline
     \end{tabular}
     \caption{\textbf{Model performance (MAE) on the external Unity dataset.} MAE: Mean absolute error.}
     \label{table:supp_unity_mae_results}
    \end{table}
    
    \begin{table}[h!]
    \centering
    \begin{tabular}{|c|c|c|c|c|c|}
    
    \hline
     Backbone & Model & Adapters & \multicolumn{3}{|c|}{Averaged landmark error} \\
     \hline
      &  &  & IVS & LVID & LVPW  \\
      &  &  & Mean $\pm$ std & Mean $\pm$ std & Mean $\pm$ std \\
     \hline
     DeepLabV3 & DeepLabV3 & \XSolidBrush & 13.86 $\pm$ 18.75 & 11.25 $\pm$ 11.12 & 9.27 $\pm$ 7.26 \\
     \hline
     \multirow{2}{*}{ViT-S}& \multirow{2}{*}{EchoApex} & \Checkmark & 6.32 $\pm$ 8.9
    & 7.66 $\pm$ 7.88 & 9.40 $\pm$ 9.62 \\
     & & \XSolidBrush & 5.98 $\pm$ 7.07 & 7.57 $\pm$ 6.49 & 9.65 $\pm$ 9.36 \\
     \hline
     \multirow{2}{*}{ViT-B}& \multirow{2}{*}{EchoApex} & \Checkmark & 6.06 $\pm$ 6.52 & 8.34 $\pm$ 7.33 & 10.28 $\pm$ 11.42 \\
     & & \XSolidBrush & 8.17 $\pm$ 12.89 & 9.03 $\pm$ 8.47 & 10.39 $\pm$ 10.68 \\
     \hline
     \end{tabular}
     \caption{\textbf{Model performance (Averaged landmark error) on the external Unity dataset.}}
     \label{table:supp_unity_lmk_results}
    \end{table}
    
    \begin{figure}[h!]
         \centering
         \begin{subfigure}{\textwidth}
             \centering
             \includegraphics[width=0.8\textwidth]{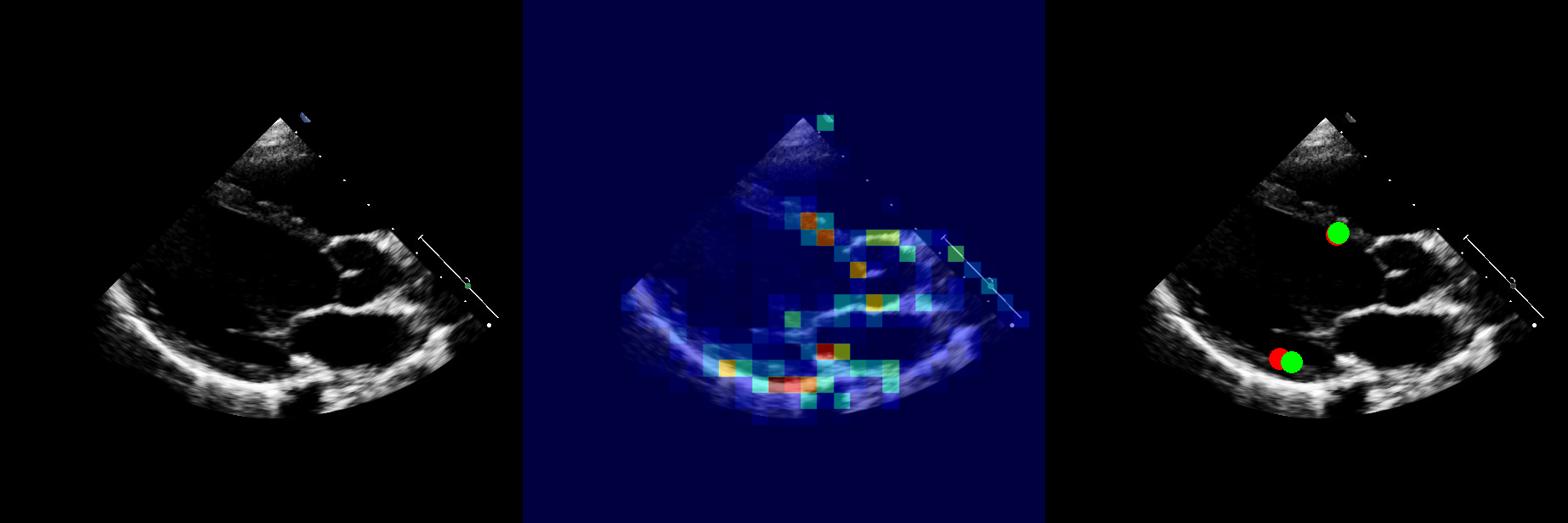}
             \caption{IVS measurement. MAE(mm): 0.12}
         \end{subfigure}
         
          \begin{subfigure}{\textwidth}
             \centering
             \includegraphics[width=0.8\textwidth]{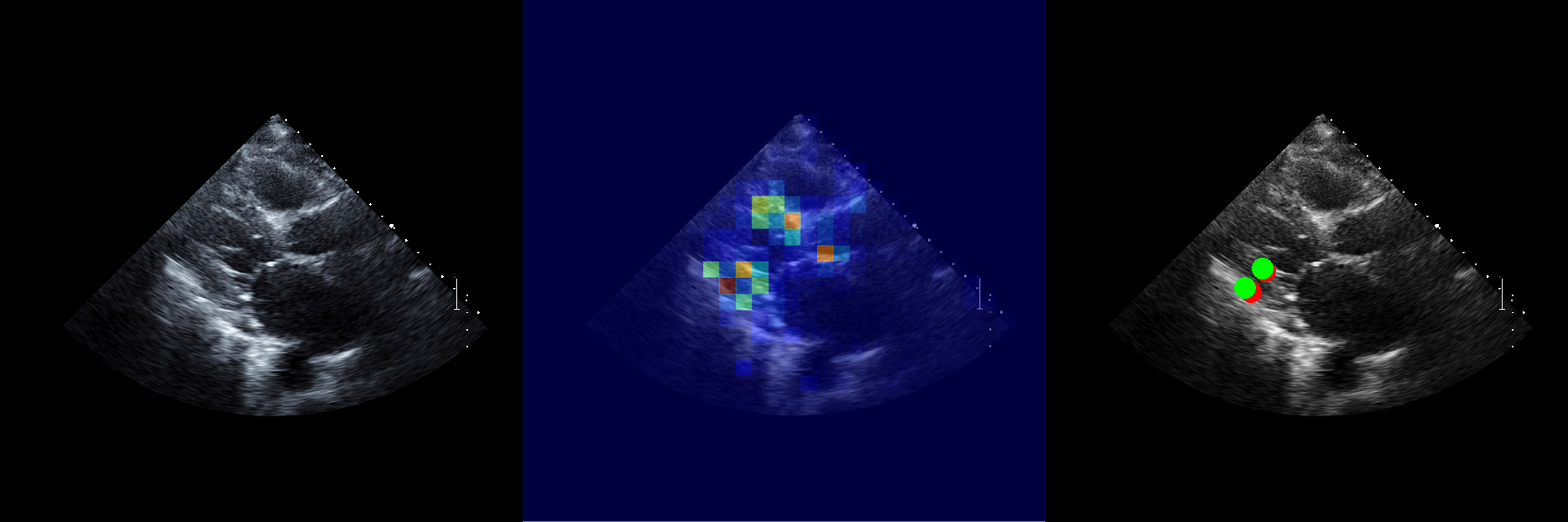}
             \caption{LVPW measurement. MAE(mm): 0.29}
         \end{subfigure}
    
         \begin{subfigure}{\textwidth}
             \centering
             \includegraphics[width=0.8\textwidth]{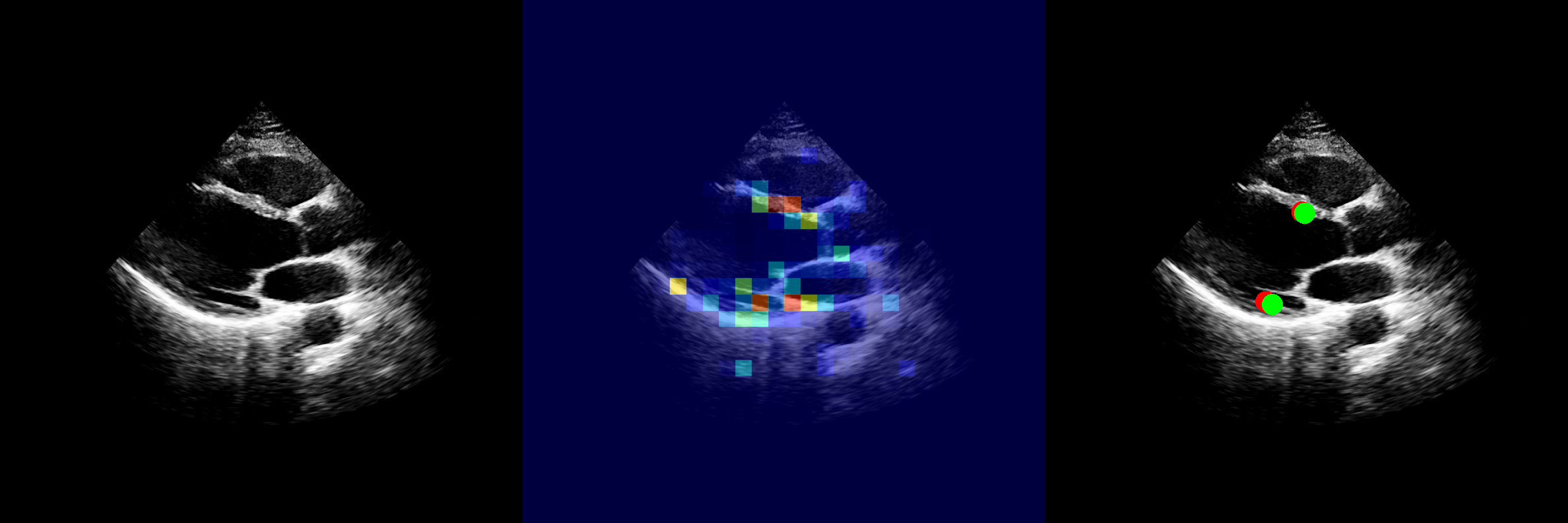}
             \caption{LVID measurement. MAE(mm): 0.29}
         \end{subfigure}
    
          \begin{subfigure}{\textwidth}
             \centering
             \includegraphics[width=0.8\textwidth]{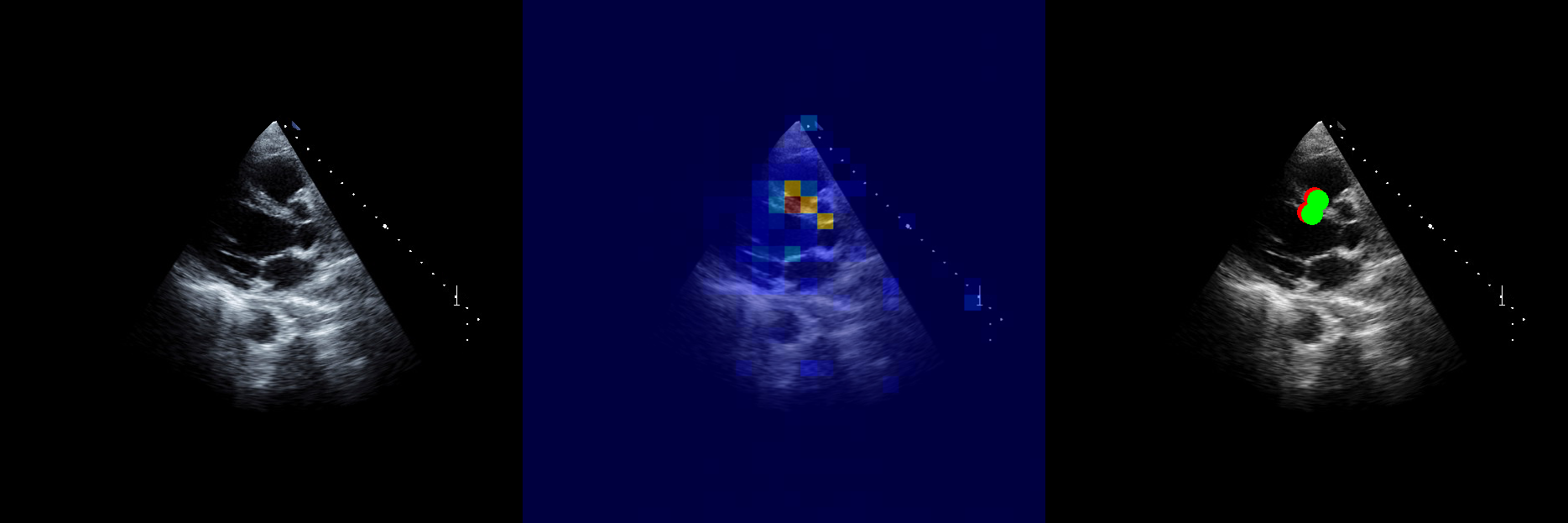}
             \caption{LVPW measurement. MAE(mm): 0.54}
         \end{subfigure}
         \caption{\textbf{Additional examples of self-attention maps from the EchoApex encoder on parasternal long-axis images for the left ventricle measurement task.} The green points indicate the ground truth human annotations while the red ones denote the model's predictions.}
         \label{fig:supp_lvh_qualitative_results}
    \end{figure}

\clearpage
\section{Supplementary Materials: Unstructured Abstract}

Quantitative evaluation of echocardiography is essential for precise assessment of cardiac condition and guiding treatment decisions. The diverse nature of echo images, including variations in probe types, manufacturers, and pathologies, poses challenges for developing artificial intelligent models that can generalize across different clinical practice. Here, we introduce EchoApex, a general-purpose vision foundation model designed for comprehensive echocardiography analysis. Pretrained on over 20 million images from 11 clinical centers, EchoApex utilizes self-supervised learning and task-specific decoders for diverse applications in echocardiography. It performs sequence view classification, interactive structure segmentation, left ventricular hypertrophy detection, and ejection fraction estimation. In benchmark evaluations, EchoApex outperforms task-specific models, achieving a mean BACC of 0.98 in classification of 18 common views, DICE of 0.93 in chamber segmentation, and a zero-shot performance improvement over specialist models in all evaluated datasets. For ejection fraction estimation, EchoApex achieves an MAE of 5.6\% and an AUC of 0.93 for cardiomyopathy detection. Despite using less than 4\% of trainable parameters with frozen encoders, EchoApex with adapter demonstrates strong performance with minimal degradation compared to fully finetuned models. This work establishes EchoApex as a scalable, general-purpose model for echocardiography, enabling efficient adaptation across various clinical tasks.

\end{appendices}

\newpage
\clearpage
\setlength\bibitemsep{3pt}
\printbibliography
\balance
\clearpage

\end{document}